\documentclass{ieeeaccess}
\usepackage{cite}
\usepackage{amsmath,amssymb,amsfonts}
\usepackage{algorithmic}

\usepackage[caption=false]{subfig}
\usepackage{caption,setspace}
\usepackage{graphicx}
\usepackage{textcomp}
\usepackage[hidelinks]{hyperref}
\usepackage[export]{adjustbox}
\usepackage{booktabs}
\usepackage[utf8]{inputenc}
\usepackage{mathtools}

\usepackage{times}
\usepackage{epsfig}

\usepackage{blindtext}

\usepackage{color,soul}
\usepackage{makecell}
\usepackage{multirow}
\usepackage{multicol}

\usepackage{bbm}
\def\BibTeX{{\rm B\kern-.05em{\sc i\kern-.025em b}\kern-.08em
    T\kern-.1667em\lower.7ex\hbox{E}\kern-.125emX}}
\begin{document}
\history{Date of publication xxxx 00, 0000, date of current version xxxx 00, 0000.}
\doi{10.1109/ACCESS.2017.DOI}

\title{SemAttNet: Towards Attention-based Semantic Aware Guided Depth Completion}

\author{\uppercase{Danish Nazir}\authorrefmark{1,2,3},
\uppercase{Alain Pagani}\authorrefmark{1},
\uppercase{Marcus Liwicki \authorrefmark{4}, Didier Stricker \authorrefmark{1,2,3}, Muhammad Zeshan Afzal \authorrefmark{1,2,3}
}
}
\address[1]{Deutsches Forschungszentrum für Künstliche Intelligenz (DFKI) Trippstadter Str. 122, 67663 Kaiserslautern}
\address[2]{Department of Computer Science, University of Kaiserslautern, 67663 Kaiserslautern, Germany}
\address[3]{Mindgrage, University of Kaiserslautern, 67663 Kaiserslautern, Germany}
\address[4]{Department of Computer Science, Luleå University of Technology, 971 87 Luleå, Sweden}

\markboth
{Author \headeretal: Preparation of Papers for IEEE TRANSACTIONS and JOURNALS}
{Author \headeretal: Preparation of Papers for IEEE TRANSACTIONS and JOURNALS}

\begin{abstract}
Depth completion involves recovering a dense depth map from a sparse map and an RGB image.~Recent approaches focus on utilizing color images as guidance images to recover depth at invalid pixels.~However, color images alone are not enough to provide the necessary semantic understanding of the scene.~Consequently, the depth completion task suffers from sudden illumination changes in RGB images (e.g., shadows).~In this paper, we propose a novel three-branch backbone comprising color-guided, semantic-guided, and depth-guided branches.~Specifically, the color-guided branch takes a sparse depth map and RGB image as an input and generates color depth which includes color cues (e.g., object boundaries) of the scene.~The predicted dense depth map of color-guided branch along-with
semantic image and sparse depth map is passed as input to semantic-guided branch for estimating semantic depth.~The depth-guided branch takes sparse, color, and semantic depths to generate the dense depth map.~The color depth, semantic depth, and guided depth are adaptively fused to produce the output of our proposed three-branch backbone.~In addition, we also propose to apply semantic-aware multi-modal attention-based fusion block (SAMMAFB) to fuse features between all three branches.~We further use CSPN++ with Atrous convolutions to refine the dense depth map produced by our three-branch backbone.~Extensive experiments show that our model achieves state-of-the-art performance in the KITTI depth completion benchmark at the time of submission.
\end{abstract}


\titlepgskip=-15pt

\maketitle

\begingroup
\renewcommand\thefootnote{}\footnotetext{© 2022 IEEE. This is the author version of a work accepted for publication in IEEE Access. The final published version is available at \href{https://doi.org/10.1109/ACCESS.2022.3214316}{https://doi.org/10.1109/ACCESS.2022.3214316}.}
\addtocounter{footnote}{-1}
\endgroup

\section{Introduction}
\label{sec:introduction}

\begin{figure}

\begin{center}
    \includegraphics[width=0.45\textwidth,keepaspectratio]{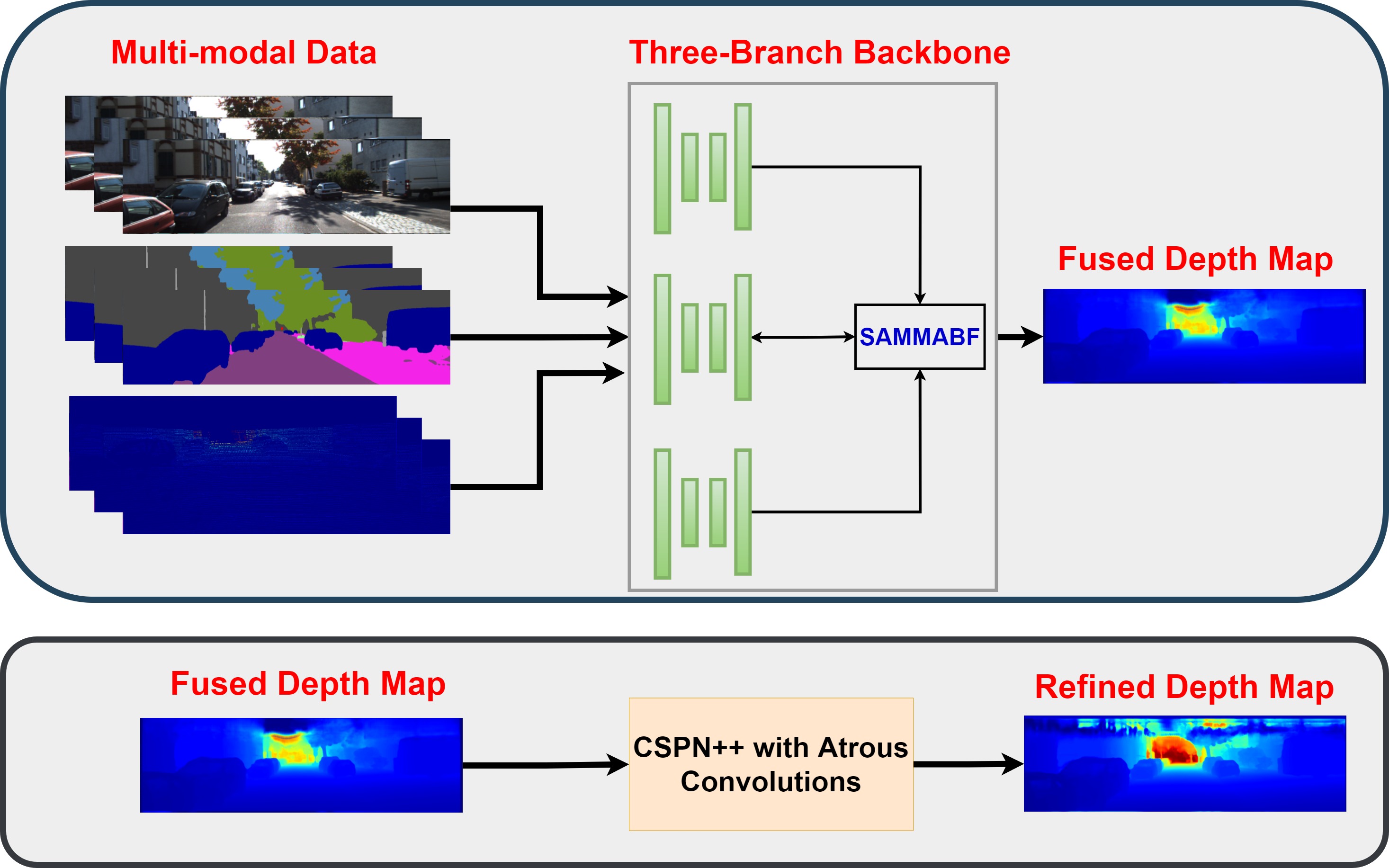} 
\end{center}
\caption{Block diagram of SemAttNet. In the first stage, the multi-modal data consisting of RGB, semantic, and LiDAR sparse depth maps are passed to our novel three-branch backbone, which outputs a fused dense depth map. Our proposed backbone utilize Semantic-aware multi-modal attention-based fusion (SAMMAFB) to adaptively fuse features between RGB, semantic and sparse depth branches.~In the second stage, the output of the three-branch backbone is further refined through CSPN++ with Atrous convolutions. }
\label{figure1}
\end{figure}
Estimation of dense depth measurements is crucial in various 3D vision and robotics applications such as augmented and mixed reality  \cite{holynski2018fast, augmented_app},  scene reconstruction  \cite{3DReconstruction, park2020non}, autonomous driving \cite{DepthNet, fisheyeautonomous}, and obstacle avoidance \cite{obstacle_avoid, tang2020learning}.~For obtaining a reliable prediction of depth in outdoor scenes, measurements from various sensors are used. Most commonly used sensors include RGB cameras in a stereo setting, light detection and ranging (LiDAR), and time of flight (ToF) cameras\cite{tof}.~Among all the sensors, LiDAR is considered the most reliable and efficient, and it produces accurate depth measurements in an outdoor setting \cite{tang2020learning}.~However, the density of LiDAR depth measurements is sparse, with a considerable amount of missing depth data\cite{KITTI, kitti_2}.~For example, the LiDAR sensor for mobile applications \cite{tang2020learning}, Velodyne HDL-64e, which is also used in the KITTI \cite{KITTI, kitti_2} dataset produces depth maps consisting of valid depth values only on $5.9 \%$ pixels.~Such sparse depth maps cannot be utilized directly in the application areas mentioned above.~Therefore, dense depth maps estimation from sparse measurements is of utmost importance. It is considered a challenging problem since the measured depth values only form $5.9 \%$ of the complete depth map.

Recent methods \cite{Qiu_2019_CVPR,hu2020PENet,yan2021rignet,lee2021depth,Multi-TaskGan,zhao2021adaptive,liu2021fcfr,cheng2020cspn++,chen2019learning}  employ the deep learning-based approaches to dense depth completion.~These methods utilize convolutional neural networks and combine sparse LiDAR data with different modalities, e.g. RGB images \cite{hu2020PENet, yan2021rignet}, semantic maps \cite{Multi-TaskGan,schneider2016semantically,chen2019towards}, and surface normal's \cite{Qiu_2019_CVPR}.~These modalities act as guidance and significantly help in the recovery of missing depth values in the sparse maps.~The idea is to actively fuse features between different modalities. Most of the current methods adopt two-branch networks architectures for feature fusion.~For instance, DeepLiDAR \cite{Qiu_2019_CVPR}, FusionNet \cite{vangansbeke2019sparse} and PENet \cite{hu2020PENet} utilize an encoder-decoder architecture to perform both early and late fusion between color images and LiDAR sparse depth maps.~The combination of early and late fusion between the modalities enhances the depth completion performance.~Recently, RigNet \cite{yan2021rignet} proposed a repetitive architecture, which repeats Hourglass network architecture for color images at multiple levels to extract features and fuse them with depth generation branch to create structure-detailed dense depth maps.~These methods rely heavily only on color images to extract color dominant information such as object boundaries to complete sparse maps.~However, color images truly do not provide this information.~It is because color images suffer from shadows and reflections, which causes irregularities in pixel values \cite{chen2019towards}.

To solve the problems in the methods mentioned earlier, we propose a novel three-branch network inspired by PENet \cite{hu2020PENet}, FusionNet \cite{vangansbeke2019sparse} and DeepLiDAR \cite{Qiu_2019_CVPR}.~It consists of color-guided (CG), semantic-guided (SG), and dense depth-guided (DG) branches.~In contrast to earlier methods \cite{Qiu_2019_CVPR,hu2020PENet,vangansbeke2019sparse}, we add SG branch to decrease the variance of depth values around object boundaries.~The CG branch produces a noisy depth estimate through the color image and LiDAR sparse depth map.~The output of the CG branch, LiDAR sparse depth and semantic map of the color image is passed to the SG branch, which actively refines the color depth and produces a depth based on semantic information. For ease of understanding, we name it semantic depth in this paper.~The semantic depth is much more reliable around object boundaries than color depth.~It is because the pixel values of semantic maps are uniform and have fewer irregularities.~Furthermore, we also use the DG branch to refine further the depth produced by the SG branch.~It takes LiDAR sparse depth map, CG and SG depth as input and produces a dense depth maps.~Similar to earlier methods \cite{hu2020PENet, zhao2021adaptive}, we also combine depth information between CG, SG and DG depths with learned confidence maps.~The confidence map weights enables us to compare the reliability of the predictions at each branch.~In the end, we perform multi-modal depth fusion between CG, SG and DG predicted dense depths to generate the final dense depth map.~The whole backbone model is trained in an end-to-end manner.~Furthermore, we apply CSPN++ with Atrous convolutions \cite{chen2017deeplab,hu2020PENet,cheng2020cspn++} on the final dense depth map predicted by the three-branch backbone for further refinement.


Since we have branches dealing with multi-modal data, we perform both early and late fusion between them for effective guidance of dense depth map generation.~Early fusion is performed by the depth-wise concatenation of color, semantic or LiDAR sparse depth maps.~The concatenation of the input is fed to the respective branch.~Late fusion is performed at the feature level and involves the fusion of multi-modal data.~The naive strategy of performing late fusion is to apply simply concatenation or addition of features maps of different modalities and it is used in many earlier works such as e.g.  \cite{tang2020learning, hu2020PENet, ma2019self}.~However, this is not an optimal way because each branch contains different information.~Therefore, we propose to apply semantic-aware multi-modal attention-based fusion block (SAMMAFB) inspired by CBAM \cite{woo2018cbam} and AFB \cite{fooladgar2019multi} to depth completion problem.~SAMMAFB applies both channel and spatial wise attention maps of the input feature maps and produces refined feature maps.~As demonstrated in Figure \ref{figure1}, we actively fuse features at multiple stages through SAMMAFB in all of our branches.

The main contributions of our work is summarized as follows:
\begin{itemize}
  \item We propose a novel three-branch backbone for sparse depth completion, which counters the sensitivity of image-guided methods to optical changes (e.g., shadows and reflections).
  \item We present a novel SAMMAFB block to actively fuse the color, semantic, and depth modalities at multiple stages in our three-branch backbone.
  \item Extensive experimental results show that our model achieves state-of-the-art results on the outdoor KITTI depth completion dataset.
  
\end{itemize}

\section{Related Work}

\subsection{Sparse Depth Based Approaches}
Earlier approaches \cite{KITTI, chodosh2018deep} based on convolutional neural networks (CNN) utilized only sparse depth maps to generate dense depth maps.~To counter the sparsity of data in sparse depth maps, Depth-Net \cite{bai2020depthnet} performed nearest neighbor interpolation in the sparse maps to fill out the holes (points with no depth values).~Later on, uncertainty aware CNN's \cite{eldesokey2020uncertainty} proposed probabilistic normalized convolutions to model the uncertainty in the sparse depth maps.~However, the obvious drawback of these approaches is that without color or semantic image guidance, the predicted depth maps lack clear object boundaries and also these methods are not suitable for real-time applications.

\subsection{Image-Guided Methods}

Image-Guided methods~\cite{park2020non,tang2020learning,hu2020PENet,yan2021rignet,zhao2021adaptive,liu2021fcfr,cheng2020cspn++,dspn,gu2021denselidar,cheng2018depth,liu2017learning} utilize multi-modal information to facilitate dense depth completion.~The multi-modal data includes RGB images, semantic images, and surface normal's, which act as reference images for generating dense depth maps.

~Methods such as SPN\cite{liu2017learning}, CSPN\cite{cheng2018depth}, CSPN++\cite{cheng2020cspn++}, DSPN\cite{dspn} and NLSPN  \cite{park2020non} focus on learning affinity matrix for high-level vision tasks including semantic segmentation and depth completion.~SPN \cite{liu2017learning} serially propagates the affinity matrix, which is inefficient for real-time systems.~CSPN \cite{cheng2018depth} improved SPN \cite{liu2017learning} by predicting affinity values of local neighbors and updating pixel values simultaneously.~Both of the methods suffer with the problem of fixed local neighborhoods, which often have irrelevant information.~To solve this problem, CSPN++ \cite{cheng2020cspn++}, DSPN \cite{dspn} and NLSPN \cite{park2020non} methods are introduced.~CSPN++ \cite{cheng2020cspn++} proposed adaptive learning of kernel sizes and the number of iterations of propagation, which helped in reducing the computation time of CSPN \cite{cheng2018depth}.~DSPN \cite{dspn} and NLSPN \cite{park2020non} learn deformable convolutional kernels to relax the fixed local neighborhood of pixels, which enabled them to focus only on relevant pixel neighbors for depth completion.~All of the methods mentioned above utilizes a single branch AutoEncoder (AE) \cite{liou2014autoencoder} network architecture.~The input of the AE is the concatenation of RGB image and sparse depth map, which outputs a dense depth map.

The two branch network architectures  \cite{Qiu_2019_CVPR,hu2020PENet,yan2021rignet,zhao2021adaptive,liu2021fcfr} consist of RGB and sparse depth map branches. The RGB branch extracts color dominant information, e.g., object boundaries, which is actively fused with a sparse depth map branch at multiple stages.
ACMNet \cite{zhao2021adaptive} introduced a symmetric gated fusion strategy for performing the fusion between two branches.~FCFR-Net proposed a channel shuffling and energy-based fusion between the two modalities.~PENET \cite{hu2020PENet}, utilized concatenation and addition operations to perform fusion at multiple levels.~Recently, RigNet \cite{yan2021rignet} proposed an efficient guidance algorithm to fuse and guide the sparse depth branch.~In addition to RGB image and sparse depth map, DeepLiDAR \cite{Qiu_2019_CVPR} introduced learning pixel-wise surface normal's of the scene, which is fused through addition with RGB branch to generate dense depth maps.~Similarly, Multi-Task GAN's \cite{Multi-TaskGan} generated semantic maps of the RGB images and concatenated them with sparse and RGB images to guide dense depth map completion.~Compared to earlier approaches, the performance of two branch methods are much better. 

Unlike existing Image-guided methods, our approach consists of a semantic-guided branch, which given the color and sparse dense depths, learns to understand the semantic meaning of the scene.~Consequently, it enables our approach to perform consistently under different lighting conditions, which is a challenging aspect of outdoor depth completion.

\subsection{Attention-based Multi-Modal Data Fusion}

Attention-based multi-modal information fusion has been studied in various computer vision applications, such as video description \cite{hori2017attention}, human motion estimation \cite{li2020attention}, emotion and sentiment classification \cite{huddar2021attention} and 
many more.~In the context of dense depth completion, the attention mechanism enables the model to focus on meaningful regions of multi-modal information, generating depths with clear and sharp structural details.~Moreover, given the sparsity of the LiDAR depth data, the attention module is important for depth completion task.~Recently, ACM Net \cite{zhao2021adaptive} proposed a Co-attention Guided Graph Propagation (CGPM), which exploits the attention mechanism in the spatial domain to propagate graphs at multiple scales.~However, spatial attention only capture local contexts within a fixed neighborhood.~Therefore, to capture both local and global attention, we propose to apply SAMMAFB between RGB, semantic, and LiDAR sparse depth modalities.~It produces both channel-wise and spatial-wise attention weights, which are applied to multi-modal information to generate refined fused feature maps.
\begin{figure*}

\begin{center}
    \includegraphics[width=\textwidth,keepaspectratio]{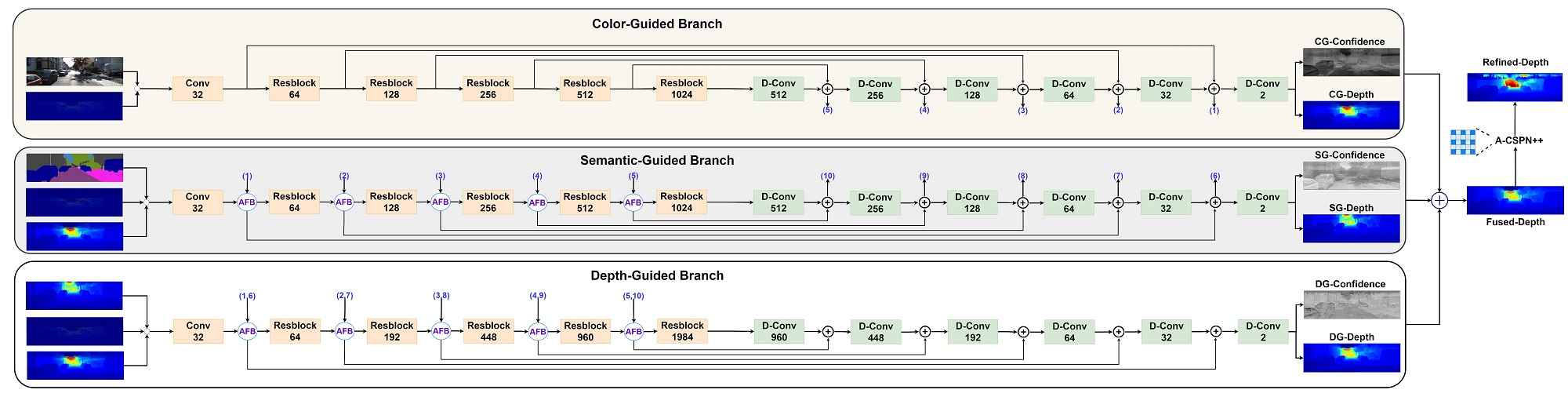} 
\end{center}
\caption{The overview of proposed SemAttNet.~It consists of a novel three-branch backbone and a CSPN++ module with Atrous convolutions.~Unlike earlier image-guided methods, we design a separate branch for learning the semantic information of the scene.~Furthermore, we propose to apply attention based fusion block (ABF) to perform semantic-aware fusion between RGB, depth, and semantic modalities.~Each branch outputs a depth map and a confidence map, which are adaptively fused to produce a fused depth map.~In the end, the fused depth map are sent to CSPN++ module with Atrous convolutions for refinement.~Note, due to shortage of space, we use AFB to represent SAMMAFB block. }

\label{SemAttNet_Architecture}
\end{figure*}

\section{Methodology}

~We formulate an image-guided depth completion problem as a two-stage task.~Figure \ref{SemAttNet_Architecture} shows the pipeline of our approach.~In the first stage, we design a three-branch backbone network to produce a dense depth map, whereas, in the second stage, we utilize CSPN++ \cite{hu2020PENet, cheng2020cspn++} with Atrous convolutions for further refinement.~The three-branch backbone consists of CG, SG and DG branches.~The purpose of the CG branch is to focus on color-dominant information, whereas the SG branch aims to learn the semantics information of the scene.~The predicted dense depths of CG and SG branches are noisy depth estimates but they contain important color and semantic cues of the scene.~The DG branch takes predicted dense depth maps of SG and CG branch along with aligned sparse map as input to produce a dense depth map.~DG branch focuses on learning depth dominant features.~All of the branches are complimentary to each other.~Therefore, we adaptively fuse them with learned confidence weights.

\subsection{Semantic-aware Multi-Modal Attention-based Fusion Block}

In order to perform fusion between RGB, semantic, and depth features, we propose to apply Semantic-aware Multi-Modal Attention-based Fusion Block (SAMMAFB) on RGB, semantic, and depth features in our three-branch backbone.~SAMMAFB helps us refine the concatenated feature maps of modalities by capturing salient feature maps while suppressing the unnecessary ones \cite{fooladgar2019multi}.~We apply SAMMAFB in two different settings, first we apply it to fuse the intermediate feature maps of RGB and semantic guided branches.~Then, for depth-guided branch, we perform the fusion between the intermediate feature maps of all branches.

\begin{figure}

\begin{center}
    \includegraphics[width=0.45\textwidth,keepaspectratio]{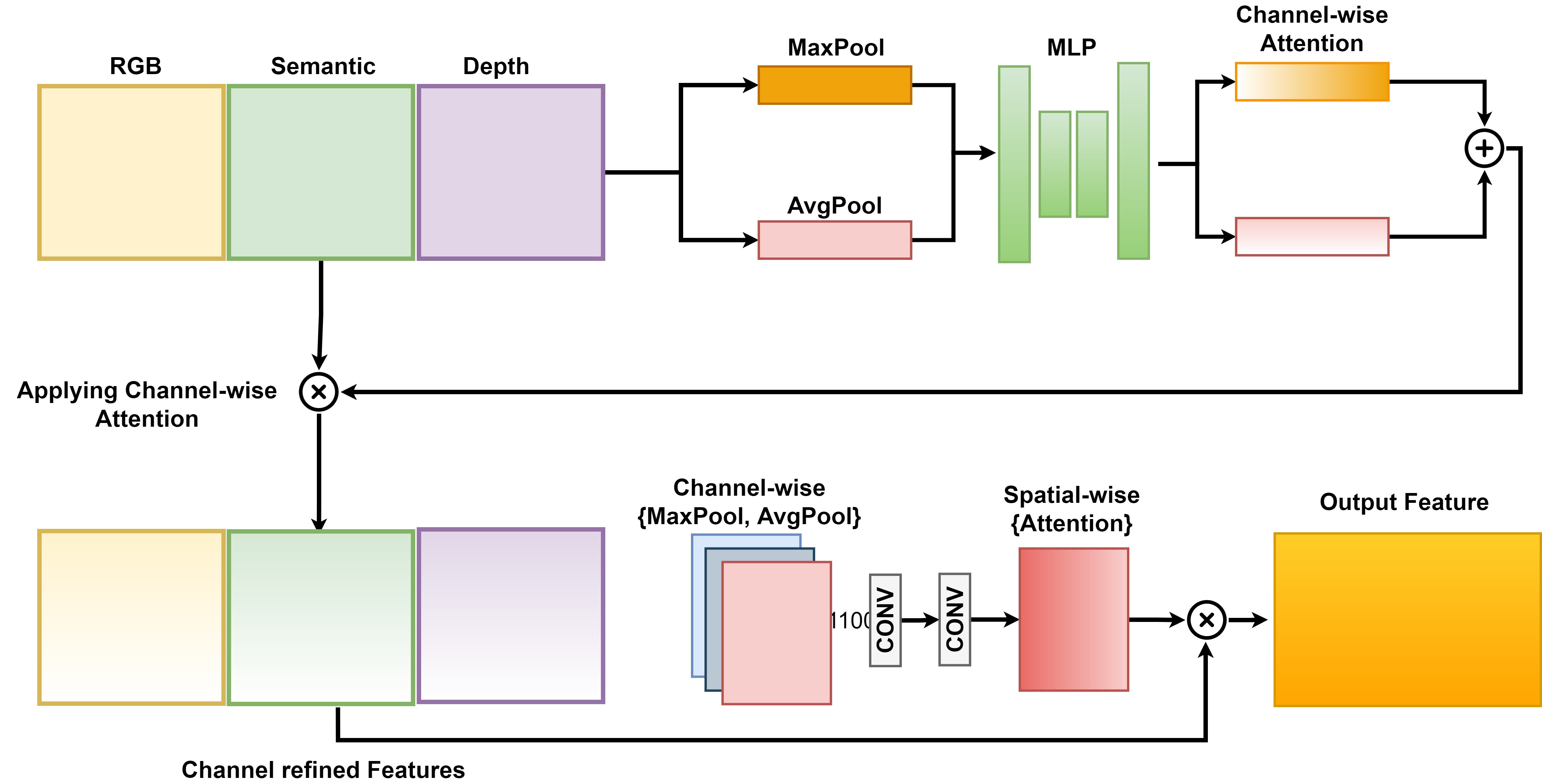} 
\end{center}
\caption{Architecture of SAMMAFB.~The input to SAMMAFB is the concatenation of RGB, semantic and sparse depth modalities.~It applies channel-wise and spatial-wise attention to the concatenated features to produce the refined fused feature maps. }

\label{SAMMAFB_Architecture}
\end{figure}

 Figure \ref{SAMMAFB_Architecture} depicts the visual representation of SAMMAFB.~It consists of a channel and spatial attention sub-modules, which enables it to capture channel-wise and spatial-wise attention.~Specifically, channel-wise attention aims to learn the important channels by assigning a weight to each channel relative to their contribution towards increasing the overall performance.~Similarly, the spatial-wise attention module focuses on learning the location of the important channels produced by the channel-wise attention sub-module.~Both of the modules complement each other and vital to producing refined fused feature maps.

Let $F_{RGB} \in \mathbb{R}^{C \times H \times W}$, $F_{Sem} \in \mathbb{R}^{C \times H \times W}$ and $F_{D} \in \mathbb{R}^{C \times H \times W}$ represent the intermediate feature maps of RGB, semantic and depth guided branches at same level, respectively and $\mathbf{F} = \left[F_{RGB};F_{Sem};F_{D}\right] \in \mathbb{R}^{3C \times H \times W}$ denotes their depth-wise concatenation.~Channel-wise attention weights are computed with Equation \ref{equation1}. 
\begin{equation}
\begin{aligned}
\mathbf{{A}_c(F)} &=\sigma(\textrm{MLP}(\textrm{AvgPool}(\mathbf{F}))+\textrm{MLP}(\textrm{MaxPool}(\mathbf{F}))) \\
&=\sigma\left(\mathbf{W_1}\left(\mathbf{W_0}\left(\mathbf{F_{avg}^{c}}\right)\right)+\mathbf{W_1}\left(\mathbf{W_0}\left(\mathbf{F_{max}^{c}}\right)\right)\right)
\end{aligned}
\label{equation1}
\end{equation}

Where $\mathbf{{A}_c(F)} \in \mathbb{R}^{C \times 1 \times 1}$ represents channel-wise attention weights of $\mathbf{F}$, $\sigma$ denotes sigmoid function, $\mathbf{W_{0}} \in \mathbb{R}^{C/r  \times C }$, $\mathbf{W_{1}} \in \mathbb{R}^{C\times C/r }$ represents the weights matrices for multi-layer perceptron (MLP) layers.~The parameter $r$ controls the number of learnable parameters in MLP layer.~$\mathbf{F_{max}^{c}}$ and $\mathbf{F_{avg}^{c}}$  denotes the max-pooled and average-pooled features respectively.

The channel-wise attentions weights are applied to $\mathbf{F}$ to obtain $\mathbf{F^{\prime}} \in \mathbb{R}^{3C \times H \times W}$ such that $\mathbf{F^{\prime}} = \mathbf{A_{c}(F)} \otimes \mathbf{F}$.~The product $\mathbf{F^{\prime}}$ is then sent to spatial-wise attention module, which is given in Equation \ref{equation_2}.
\begin{equation}
\mathbf{A_{s}}\left(\mathbf{F^{\prime}}\right)=\sigma\left(\textrm{Conv}\left(\left[\mathbf{F_{avg}^{c}} ; \mathbf{F_{max}^{c}}\right]\right)\right)
\label{equation_2}
\end{equation}
Where $\mathbf{{A}_s(F)} \in \mathbb{R}^{1 \times H \times W}$ represents spatial-wise attention weights of $\mathbf{F}$, $\sigma$ denotes sigmoid function.~$\mathbf{F_{max}^{c}}$ and $\mathbf{F_{avg}^{c}}$  denotes the max-pooled and average-pooled features respectively. 

The spatial-wise attentions weights are applied to $\mathbf{F^{\prime}}$ to obtain final output of SAMMAFB i.e. refined fused feature maps $\mathbf{F^{\prime\prime}} \in \mathbb{R}^{3C \times H \times W}$ such that $\mathbf{F^{\prime\prime}} = \mathbf{A_{s}}(\mathbf{F^{\prime}}) \otimes \mathbf{F^{\prime}}$.

\subsection{The Three-branch Backbone}
The earlier two-branch methods \cite{hu2020PENet,yan2021rignet,zhao2021adaptive,liu2021fcfr} relied heavily on the color image to extract both semantic and color-dominant information.~However, color images alone might not be able to provide this information.~Therefore, to overcome this limitation, we propose a three-branch backbone consisting of CG, SG and DG branches.~The addition of the SG branch enables better learning of semantic cues of the scene, which help predict depth maps with accurate and sharp structure in the DG branch.~The dataset comprises of color and semantic images that are represented by $C = \mathbb{R}^{N \times 3 \times H \times W}$ and $S=\mathbb{R}^{N \times 3 \times H \times W}$, whereas $ D=\mathbb{R}^{N \times 1 \times H \times W}$ and $D^{gt} = \mathbb{R}^{N \times 1 \times H \times W}$ denotes sparse and ground truth depth maps, and $N$ denotes the number of samples.~Note, that we don't have any supervision of confidence maps due the unavailability of ground truth.~Therefore, confidence maps are trained indirectly based on the $\ell_2$ loss of each branch.

\subsubsection{Color-guided Branch}
The CG branch aims to learn important color cues for dense depth completion.~It takes a color image concatenated with an aligned sparse depth map as input and outputs a dense depth map.~The concatenation of an aligned sparse depth map with a color image helps in the prediction of a dense depth map \cite{hu2020PENet}.~The CG branch follows an encoder-decoder network architecture with skip connections.~The encoder consists of ten ResNet \cite{he2016deep} blocks, whereas the decoder consists of one convolution and five transpose convolution layers for upsampling.~The output of this branch consist of dense depth map and a confidence map.~The resultant dense depth map of this branch is a noisy depth estimate, but it provides a baseline for learning structural information of the scene in other branches.

Let $ \mathbf{X_{cg}} =  \left[C_B;D_B \right] \in \mathbb{R}^{B \times 4 \times H \times W}$ represent the concatenation of batch of size $B$ of color images $ C_B \in C$ and sparse depth maps $D_B \in C$.~The CG branch takes $\mathbf{X_{cg}}$ as input, and in the first step, it is encoded to a hidden representation $\mathbf{\phi_{cg}} \in \mathbb{R}^{B \times 1024 \times H \times W}$ based on Equation \ref{equation_CD_1}.~Then, $\mathbf{\phi_{cg}}$ is decoded into a confidence map $C_{cg} \in  \mathbb{R}^{B \times 1 \times H \times W}$  and a color depth map $D_{cg} \in  \mathbb{R}^{B \times 1 \times H \times W}$ based on Equation \ref{equation_CD_2}. 
\begin{equation}
    \begin{aligned}
            \mathbf{\phi_{cg}} =   \textrm{f}(\mathbf{W_{cg}}\mathbf{X_{cg}} + b_{cg})
    \end{aligned}
    \label{equation_CD_1}
\end{equation}
\begin{equation}
    \begin{aligned}
            D_{cg}, C_{cg} =   \textrm{g}(\mathbf{V_{cg}}\mathbf{\phi_{cg}} + c_{cg})
    \end{aligned}
    \label{equation_CD_2}
\end{equation}

Where $\mathbf{W_{cg}}$ and $\mathbf{V_{cg}}$ represents encoder and decoder weight matrices.~The variables  $b_{cg}$ and $c_{cg}$ denote encoding and decoding bias values.~$\textrm{f(.)}$ and $\textrm{g(.)}$ represent activation functions.
\begin{equation}
    \begin{aligned}
      L_{cg} =   \textrm{argmin}_{D_{cg}} \ || D ^{gt} - D_{cg} ||^2
    \end{aligned}
    \label{loss_CD}
\end{equation}
Equation \ref{loss_CD} defines the $\ell_2$ loss for CG branch.~Since the ground truth contains invalid depth values, we only consider pixels with valid depth values for computing loss.

\subsubsection{Semantic-guided Branch}
Semantic cues help to understand the scene and are essential for depth completion task \cite{semantic}.~However, CG branch alone is not enough to learn semantic information.~Therefore, we propose an SG branch to encourage learning effective semantic cues and to complement RGB images for depth completion.~The SG branch takes the concatenation of color depth, semantic image and sparse depth map as an input and outputs a dense depth map, which consist of both color and semantic cues.~The KITTI depth completion dataset does not provide aligned semantic maps of RGB images.~Therefore, we transform the RGB images to semantic maps through a pre-trained WideResNet38 \cite{wu2016wider} model on KITTI semantic segmentation benchmark \cite{Alhaija2018IJCV}.~Furthermore, Inspired by earlier works \cite{tang2020learning,hu2020PENet}, we fuse the decoder features from the CG branch into the corresponding encoder features of the SG branch.~The features from CG and SG are sent to SAMMAFB, which outputs refined fused depth feature maps.


Similar to CG branch, we define $ \mathbf{X_{sg}}=\left[D_{cg}^B;S_{B};D_{B} \right] \in \mathbb{R}^{B \times 4 \times H \times W}$ to represent the concatenation of batch size $B$ of color depths $D_{cg}^B \in D_{cg}$, color images $C_B \in C$ and sparse depth maps $D_B \in D$.~The color depth consists of important color cues e.g. object boundaries and its concatenation will complement in learning semantic cues.~The matrix $\mathbf{X_{sg}}$ is passed as an input to SG branch.~In the first step, the input is encoded to a hidden representation $\phi_{sg}  \in \mathbb{R}^{B \times 1024 \times H \times W}$ based on Equation \ref{equation_SD_1}.~In the second step, the encoded representation $\phi_{sg}$ is decoded into a confidence map $C_{sg}  \in \mathbb{R}^{B \times 1 \times H \times W}$ and a semantic depth map $D_{sg}  \in \mathbb{R}^{B \times 1 \times H \times W}$ based on Equation \ref{equation_SD_2}. 
\begin{equation}
    \begin{aligned}
            \phi_{sg} =   \textrm{f}\mathbf{(W_{sg}^TX_{sg}} + b_{sg})
    \end{aligned}
    \label{equation_SD_1}
\end{equation}
\begin{equation}
    \begin{aligned}
            D_{sg}, C_{sg} =   \textrm{g}\mathbf{(V_{sg}^T\phi_{sg}} + c_{sg})
    \end{aligned}
    \label{equation_SD_2}
\end{equation}

Where $\mathbf{W_{sg}}$ and $\mathbf{V_{sg}}$ represents encoder and decoder weight matrices.~The variables $b_{sg}$ and $c_{sg}$ represents encoding and decoding bias values.~$\textrm{f(.)}$ and $\textrm{g(.)}$ denotes activation functions.
\begin{equation}
    \begin{aligned}
            L_{sg} =   \textrm{argmin}_{D_{sg}} \ || \ D ^{gt} -D_{sg} ||^2
    \end{aligned}
    \label{loss_SG}
\end{equation}
~Equation \ref{loss_SG} defines the $\ell_2$ loss for SG branch.~we only consider pixels with valid depth values for loss calculation because the ground truth contains invalid depth values.

\subsubsection{Depth-guided Branch}

The DG branch focuses on learning depth-dominant features, which help generate accurate dense depth maps.~It takes the concatenation of CG, SG, and sparse depth as input and outputs a dense depth map.~Similar to the feature fusion strategy in the CG and SG branch, we fuse decoder features from CG and SG branches into the corresponding encoder features of the DG branch.~However, in DG branch, SAMMAFB take feature maps from three modalities as input and outputs refined fused feature maps.~The refined fused feature maps consists of useful information from CG and SG branches, which guides the DG branch in learning effective depth feature representations.~The network architecture is similar to CG and SG branches, but we add two more layers in encoder and decoder architectures to accommodate CG and SG branch decoder features.

~We define $\mathbf{X_{dg}} =  \left[D_{cg}^{B};D_{sg}^{B};D_B \right]  \in \mathbb{R}^{B \times 3 \times H \times W}$ to represent the concatenation of batch of size $B$ of color depths $D_{cg}^B$, semantic depths $D_{sg}^B$ and sparse depth maps $D_B$.~The DD branch takes $\mathbf{X_{dg}}$ as an input and encodes it to a hidden representation $\mathbf{\phi_{dg} }  \in \mathbb{R}^{B \times 1984 \times H \times W}$ based on Equation \ref{equation_DD_1}.~Afterwards, the encoded representation $\mathbf{\phi_{dg}}$ is decoded into a confidence map $C_{dg}  \in \mathbb{R}^{B \times 1 \times H \times W}$ and a dense depth map $D_{dg}  \in \mathbb{R}^{B \times 1 \times H \times W}$ based on Equation \ref{equation_DD_2}. 
\begin{equation}
    \begin{aligned}
        \mathbf{\phi_{dg}} =   \textrm{f}\mathbf{(W_{dg}^TX_{dg}} + b_{dg})
    \end{aligned}
    \label{equation_DD_1}
\end{equation}
\begin{equation}
    \begin{aligned}
            D_{dg}, C_{dg} =   \textrm{g}\mathbf{(V_{dg}^T\phi_{dg}} + c_{dg})
    \end{aligned}
    \label{equation_DD_2}
\end{equation}

Where $\mathbf{W_{dg}}$ and $\mathbf{V_{dg}}$ represents encoder and decoder weight matrices.~The variables $b_{sg}$ and $c_{sg}$ represents encoding and decoding bias values.~$\textrm{f(.)}$ and $\textrm{g(.)}$ denotes activation functions.
\begin{equation}
    \begin{aligned}
            L_{dg} =   \textrm{argmin}_{D_{dg}} \ || \ D ^{gt} - D_{dg} ||^2
    \end{aligned}
    \label{loss_SD}
\end{equation}
~Equation \ref{loss_SD} defines the $\ell_2$ loss for DD branch.~Similar to SG and CG branches, we only consider pixels with valid depth values in ground truth for loss calculation.

\subsubsection{Multi-modal Depth Fusion}
~Similar to earlier approaches \cite{hu2020PENet, vangansbeke2019sparse}, we fuse the predicted depth maps of each branch by using learned confidence weights.~Equation \ref{equation_fusion} shows the mathematical representation of our depth map fusion strategy.
 \begin{equation}
     \begin{aligned}
             D_f =  \frac{ e^{C_{cg}} D_{cg} + e^{C_{sg}}D_{sg} + e^{C_{dg}} D_{dg}  } { e^{C_{cg}} + e^{C_{sg}} + e^{C_{dg}} } 
     \end{aligned}
     \label{equation_fusion}
 \end{equation}

Where $D_f$ denotes the fused depth map, $D_{cg}$, $D_{sg}$ and $D_{dg}$ represents the depth maps predicted by CG, SG and DG branches and $e^{C_{cg}}$, $e^{C_{sg}}$, $e^{C_{dg}}$ denotes the learned confidence maps of CG, SG and DG predicted depth maps.

Following the CG, SG and DG branches, we also calculate the loss on fused depth map by taking $\ell_2$ norm between the fused depth map and ground truth as depicted in Equation \ref{loss_DF}.~We only consider pixels with valid depth values for loss calculation of fused depth map. 
\begin{equation}
    \begin{aligned}
            L_{fused} =   \textrm{argmin}_{D_{f}} \ || \ D ^{gt} - D_{f} ||^2
    \end{aligned}
    \label{loss_DF}
\end{equation}

\subsubsection{Loss Function}
The training loss of our three-branch backbone consists of the sum of CD, SD, DD and fused depth losses.~Equation \ref{loss_sum} defines the training loss of our three-branch backbone.
\begin{equation}
    \begin{aligned}
            L_{total} =  \lambda_{cg}L_{cg} + \lambda_{sg}L_{sg} + \lambda_{dg}L_{dg} + L_{fused}   
    \end{aligned}
    \label{loss_sum}
\end{equation}

Where $\lambda_{cg}, \ \lambda_{sg}, \ \lambda_{dg}$ denotes the weights for color-guided, semantic-guided and depth-guided branches.


\subsection{CSPN++ with Atrous Convolutions}

The reason for applying CSPN++ for refining the dense depth maps produced by our three-branch backbone is two folds.~First, it recovers depth values at valid pixels.~Second, enabling a smooth transition between the neighbours of our completed dense depth map.~Furthermore, inspired by PENet \cite{hu2020PENet}, we use Atrous convolutions \cite{yu2016multiscale} to enlarge the propagation neighborhood of pixels \cite{xu2020deformable}.

Given a fused depth map $D_{f} \in \mathbb{R}^{h \times w \times 1}$, it is embedded into a hidden representation $H \in \mathbb{R}^{h \times w \times c}$, the one step propagation of CSPN++ with Atrous convolution is given in Equation \ref{cspn++_1}.
\begin{equation}
\begin{gathered}
    H_{t+1}(x_{i}) =\kappa_{x_{i}}\left(x_{i}\right) \odot H_{t}\left(x_{i}\right)+\sum_{x_{j} \in \mathcal{N}_{k}(x_i)}\kappa_{x_{i}}(x_{j}) \odot H_{t}(x_{j})\\
    \kappa_{x_{i}}\left(x_{j}\right)=\frac{\hat{\kappa}_{x_{i}}\left(x_{j}\right)}{\sum_{x_{j} \in \mathcal{N}}\left|\hat{\kappa}_{x_{i}}\left(x_{j}\right)\right|}, \kappa_{x_{i}}\left(x_{i}\right)=1-\sum_{x_{j} \in \mathcal{N}} \kappa_{x_{i}}\left(x_{j}\right)
\end{gathered}
\label{cspn++_1}
\end{equation}

Where $\mathcal{N}_{k}\left(x\right) = \left\{x_{n}+ l \mid l \in \mathbb{R} \right\}$ denotes the neighborhood pixels of $x_{i}$, $\hat{\kappa_{x_{i}}} \in \mathbb{R }^ {k \times k \times c}$ represents the affinity values produced by the network.~During propagation, pixel $x_{i}$ receives information from enlarged neighbourhood $\mathcal{N}_{k}(x_i)$.

\section{Experiments}

\begin{figure*} [ht!]
  \centering
	\subfloat[RGB Image]{
     \includegraphics[width=0.25\textwidth]{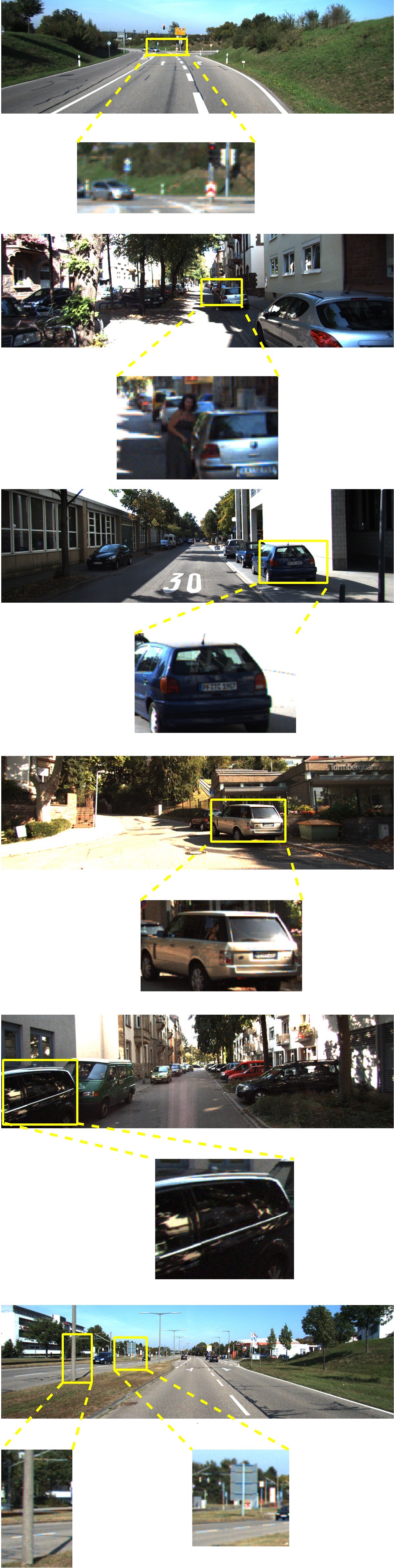}}
  \label{1a}\hspace*{-.05in}
	\subfloat[PENET]{
      \includegraphics[width=0.25\textwidth]{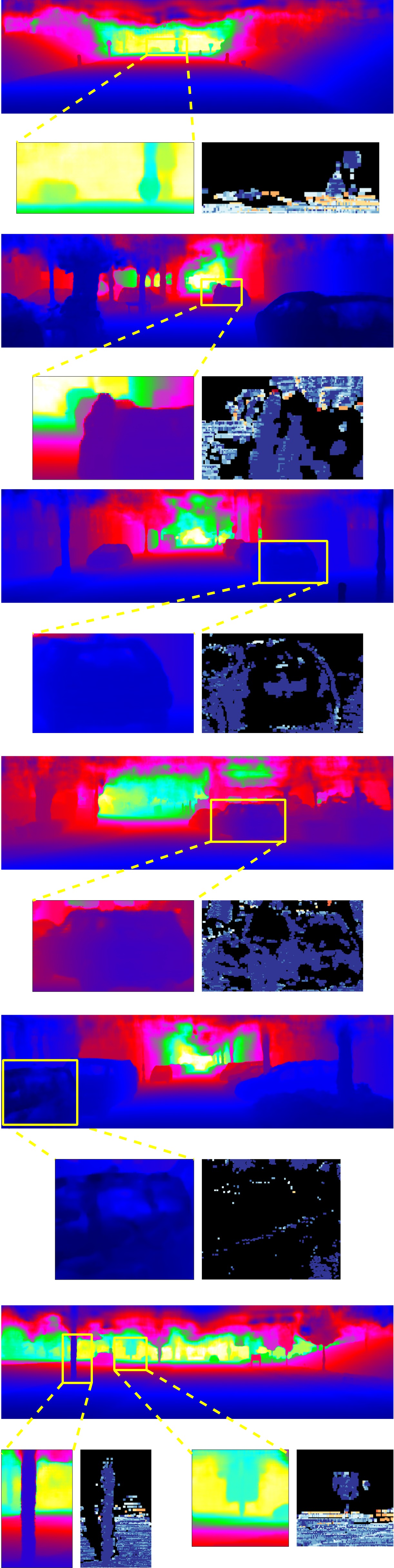}}
  \label{1b}\hspace*{-.05in}
	\subfloat[RigNet]{
      \includegraphics[width=0.25\textwidth]{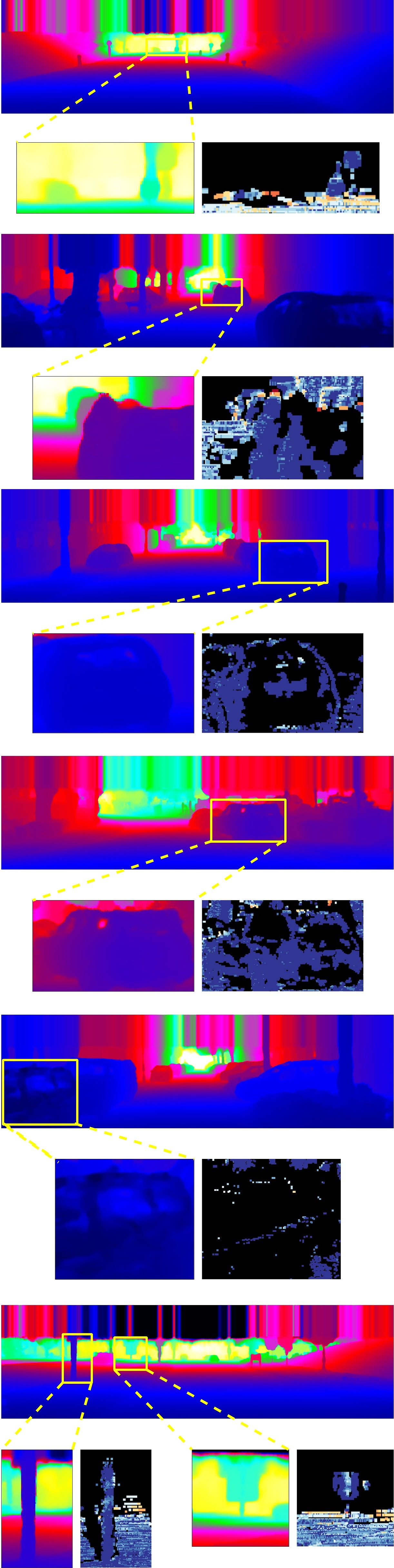}}
  \label{1c}\hspace*{-.05in}
	\subfloat[SemAttNet(ours)]{
      \includegraphics[width=0.25\textwidth]{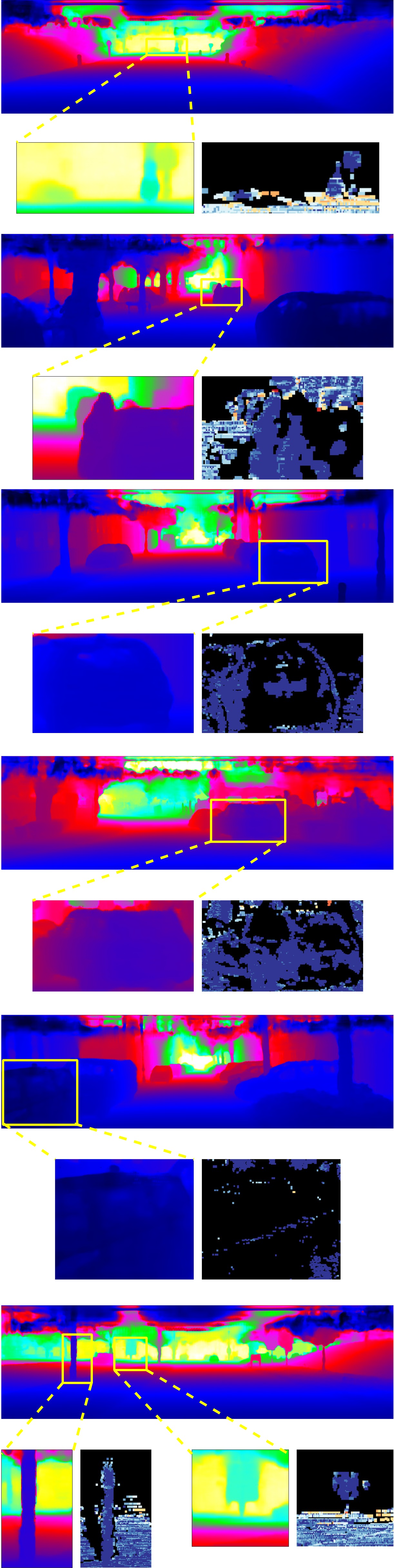}}
   \label{1d} 
\caption{Qualitative results on KITTI depth completion test set.~The results are obtained by online KITTI depth completion leaderboard.~From left-to-right (a) RGB Image, (b) PENet \cite{hu2020PENet}, (c) RigNet \cite{yan2021rignet}, (d) SemAttNet(ours).~The correct estimates in errors maps are given in blue color, whereas wrong estimate are depicted by warmer colors.}
 \label{qualitative_results}
\end{figure*}

\begin{table}

\caption{Quantitative comparison with state-of-the-art results on KITTI depth completion benchmark, ranked by \textbf{RMSE}.~The results of other methods are obtained from KITTI leaderboard.~Our method i.e. SemAttNet achieves state-of-the-art performance at the time of submission.}
\label{table1}
\begin{tabular}{l|cccc}
\hline
Method & \makecell{RMSE \\ mm} & \makecell{MAE \\ mm} & \makecell{iRMSE \\ 1/km} & \makecell{iMAE \\ 1/km}\\
\hline
TWISE \cite{imran2021depth} & 840.20 & \textbf{195.58} & 2.08 & \textbf{0.82}  \\
DSPN \cite{dspn} & 766.74 & 220.36 & 2.47 & 1.03  \\
DLiDAR \cite{Qiu_2019_CVPR} & 758.38 & 226.50 & 2.56 & 1.15  \\
FuseNet \cite{chen2019learning} & 752.88 & 221.19 & 2.34 & 1.14  \\
ACMNet \cite{zhao2021adaptive} & 744.91 & 206.09 & 2.08 & 0.90  \\
CSPN++ \cite{cheng2020cspn++} & 743.69 & 209.28 & 2.07 & 0.90  \\
NLSPN \cite{park2020non} & 741.68 & 199.59 & \textbf{1.99} & 0.84  \\
GuideNet \cite{tang2020learning} & 736.24 & 218.83 & 2.25 & 0.99  \\
FCFRNet \cite{liu2021fcfr} & 735.81 & 217.15 & 2.20 & 0.98  \\
PENet \cite{hu2020PENet} & 730.08 & 210.55 & 2.17 & 0.94  \\
RigNet \cite{yan2021rignet} & 713.44 & 204.55 & 2.16 & 0.92  \\
\hline
\textbf{SemAttNet} & \textbf{709.41} & 205.49 & 2.03 & 0.90  \\
\hline
\end{tabular}

\vspace{-11pt}
\end{table}
\subsection{Dataset}

We evaluate the performance of our method against different state-of-the-art (SoTA) methods on KITTI depth completion benchmark \cite{KITTI, kitti_2}.~The KITTI dataset is a large outdoor dataset for autonomous vehicles.~It provides 85K sparse depth maps with corresponding RGB images for training, 7K for validation, and 1K for testing.~The sparse depth map in the KITTI dataset is generated by using LiDAR HDL-64, which provides valid depth values on only $5.9 \%$ of all pixels \cite{KITTI, kitti_2}.~However, the ground-truth contains valid depth values on $16 \%$ of all the pixels.~The ground-truth is generated by accumulating LiDAR and stereo estimation of the scenes \cite{KITTI, kitti_2}.~For all of our experiments, we use the official validation set, which consists of 1K frames  \cite{KITTI, kitti_2}.~During training, we bottom crop the semantic images, RGB images, and sparse depth map from $375 \times 1275$ to $352 \times 1252$. 

\subsection{Evaluation Metrics}
We use the official evaluation metrics of the KITTI depth completion benchmark \cite{KITTI, kitti_2} to evaluate the performance of our approach.~The official evaluation measures consist of root mean squared error (RMSE), mean absolute error (MAE), root mean squared error of inverse depth (iRMSE), and mean absolute error of the inverse depth (iMAE).~Among the evaluation metrics, RMSE is chosen to rank the submissions on the KITTI leaderboard. 
\begin{table*}
\caption{Ablation study on KITTI depth completion selected validation dataset.~CG, SG and DG stands for color-guided, semantic-guided and depth-guided branches and concat, add refers to simple concatenation and addition of multi-modal features.}
\label{table2}
\begin{center}

\begin{tabular}{c|cc|ccc|c|cccc}

\hline
\multirow{2}{*}{Methods} & 
\multicolumn{2}{c|}{ Branches} & 
\multicolumn{3}{c|}{ Fusion} &
\multirow{2}{*}{\makecell{CSPN++ \\ Refinement}} & 
\multirow{2}{*}{ \makecell{RMSE \\ mm}} 
\\
\cline{2-6}&  CG + DG &  CG + SG + DG  &  add & concat & SAMMAFB & & \\
\hline

(a)  & \checkmark & & \checkmark &  &     &   &  782.89    \\
(b)  & \checkmark & & & \checkmark &     &   &  781.66     \\
(c)  & \checkmark & & & \checkmark &     &  \checkmark &  762.84    \\
\hline
(d)  &  & \checkmark &   & \checkmark &     &   &  755.16     \\
(e)  &  & \checkmark &   & \checkmark &     &  \checkmark & 750.06     \\

(f)  &  & \checkmark &   &  & \checkmark    &   &  753.02    \\
(g)  &  & \checkmark &   &  & \checkmark    &  \checkmark & \textbf{738.13}     \\
\hline

\end{tabular}

\end{center}

\end{table*}

\subsection{Implementation Details}
\begin{figure*}[ht!]

\fontsize{8}{9}\selectfont 
\begin{tabular}{c @{\hspace{0.7\tabcolsep}} c @{\hspace{0.4\tabcolsep}} c  @{\hspace{0.4\tabcolsep}} c}

\vspace{2mm}    
\rotatebox[origin=c]{90}{RGB}
&
\includegraphics[valign=m,width=0.33\textwidth]{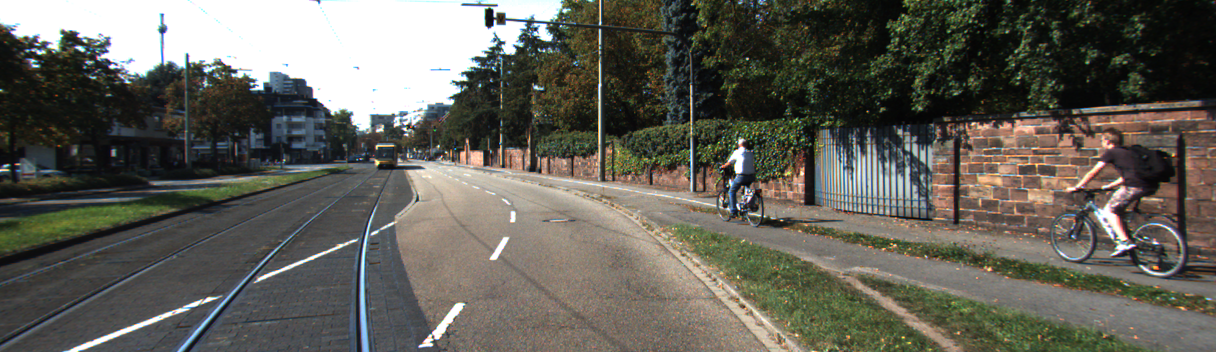}

&
\includegraphics[valign=m,width=0.33\textwidth]{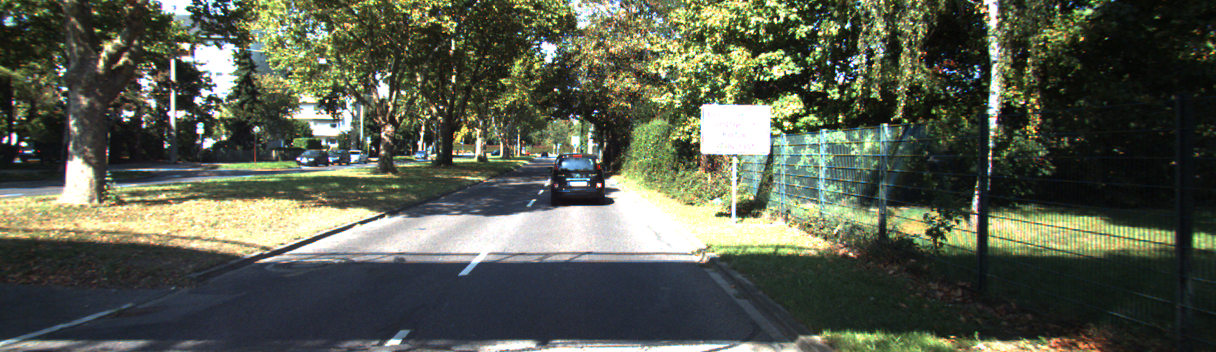}

&
\includegraphics[valign=m,width=0.33\textwidth]{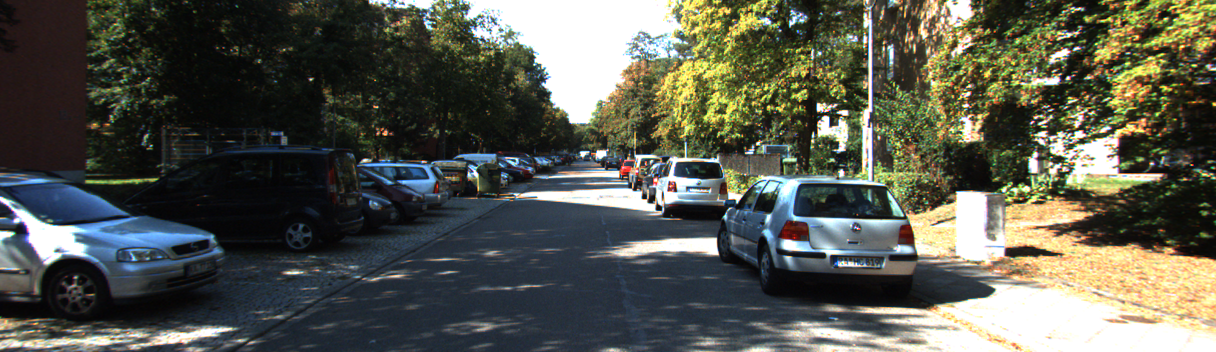}\\
\vspace{2mm}
\rotatebox[origin=c]{90}{Sem.}
&
\includegraphics[valign=m,width=0.33\textwidth]{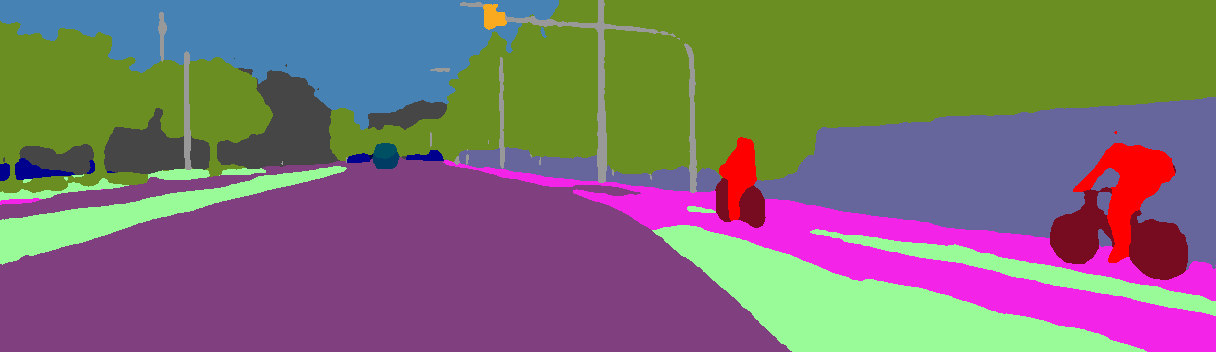}

&
\includegraphics[valign=m,width=0.33\textwidth]{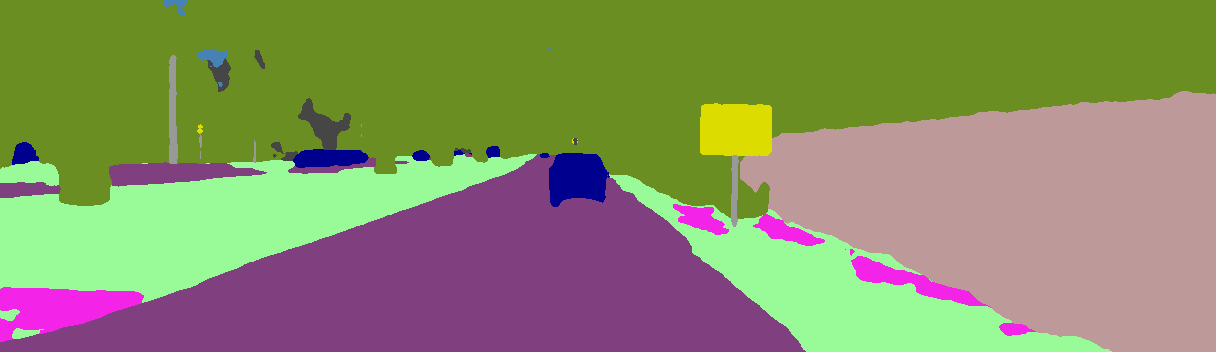}

&
\includegraphics[valign=m,width=0.33\textwidth]{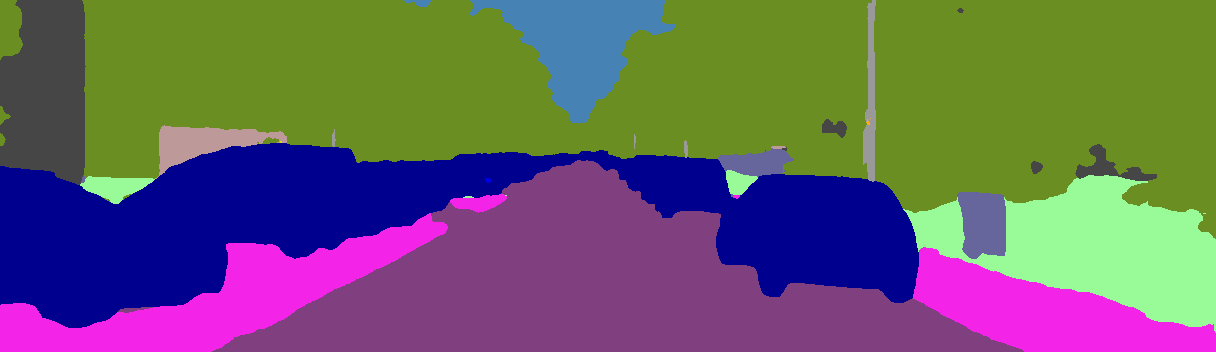}
\\
\vspace{2mm}

\rotatebox[origin=c]{90}{SD}
&
\includegraphics[valign=m,width=0.33\textwidth]{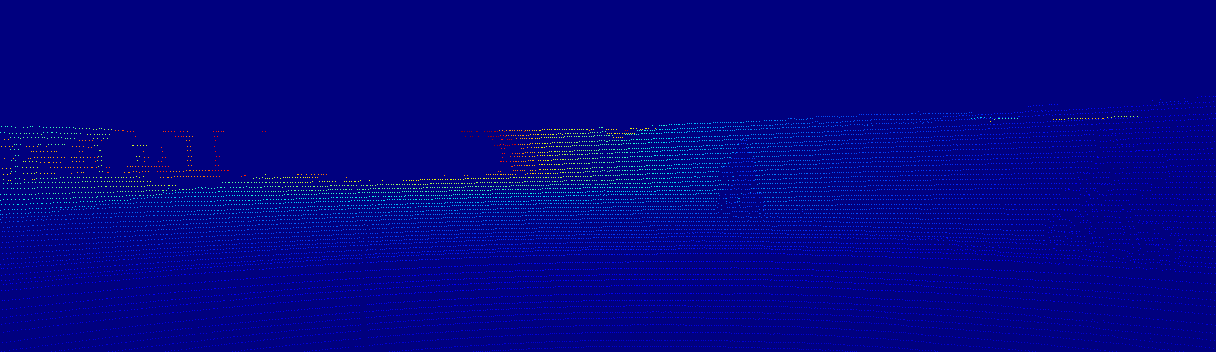}

&
\includegraphics[valign=m,width=0.33\textwidth]{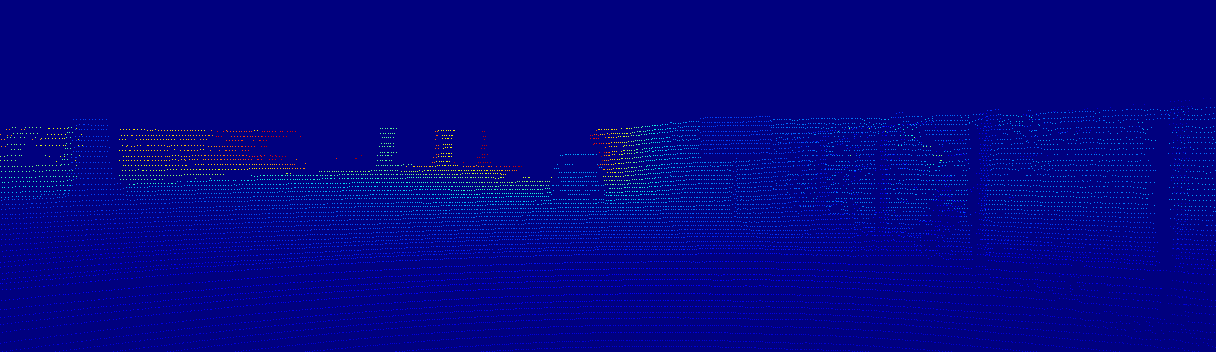}

&
\includegraphics[valign=m,width=0.33\textwidth]{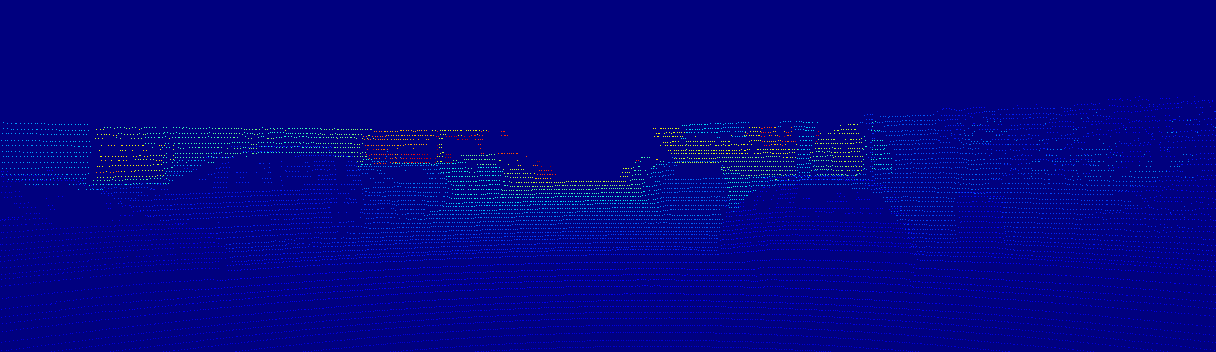}
\\
\vspace{2mm}
\rotatebox[origin=c]{90}{GT}
&
\includegraphics[valign=m,width=0.33\textwidth]{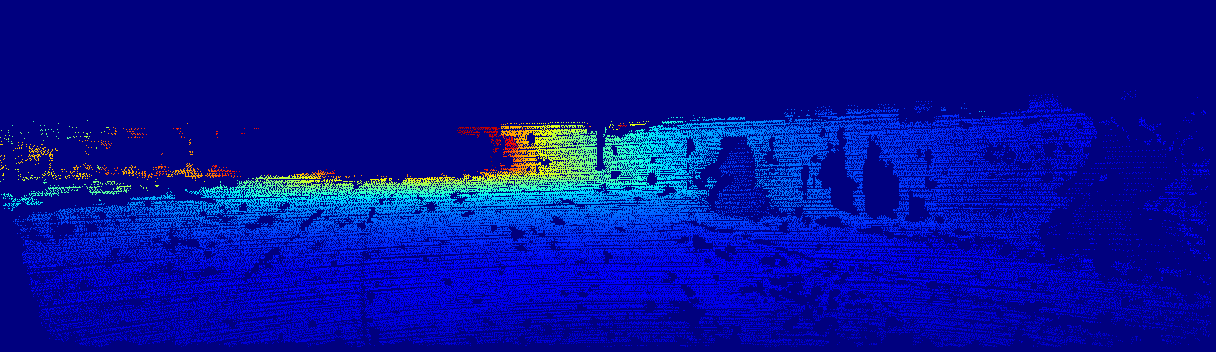}

&
\includegraphics[valign=m,width=0.33\textwidth]{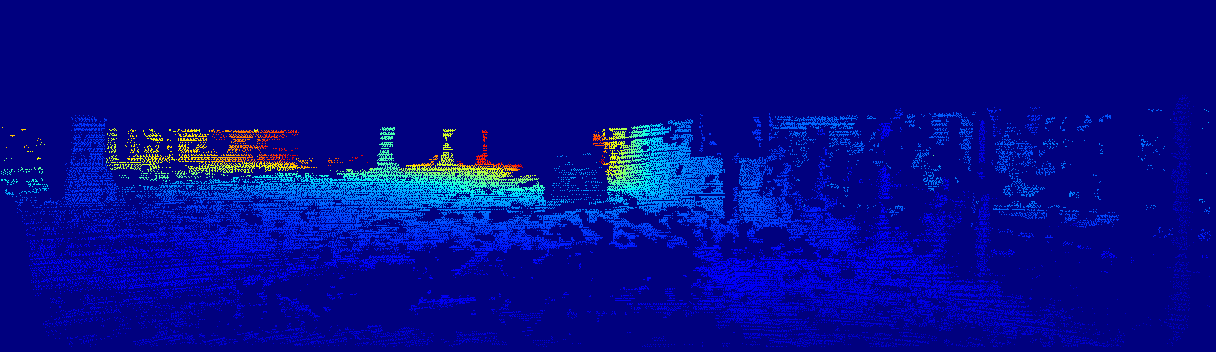}

&
\includegraphics[valign=m,width=0.33\textwidth]{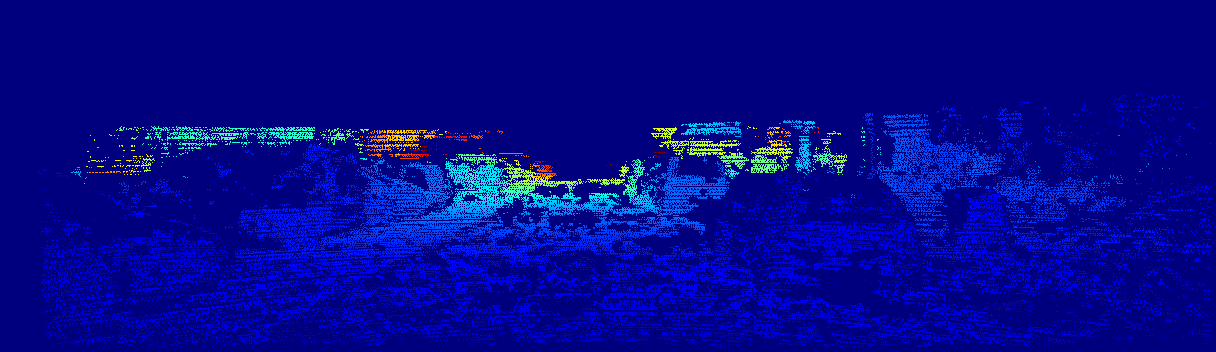}
\\
\vspace{2mm}

\rotatebox[origin=c]{90}{CG-Dep.}
&
\includegraphics[valign=m,width=0.33\textwidth]{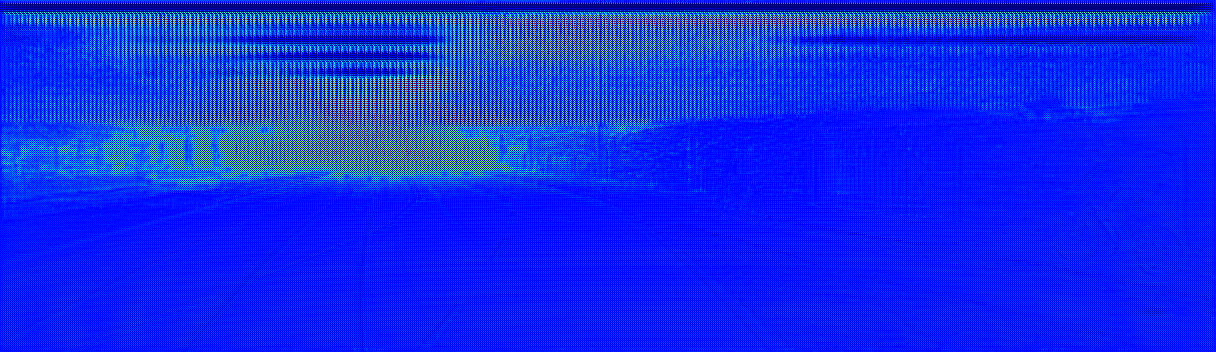}

&
\includegraphics[valign=m,width=0.33\textwidth]{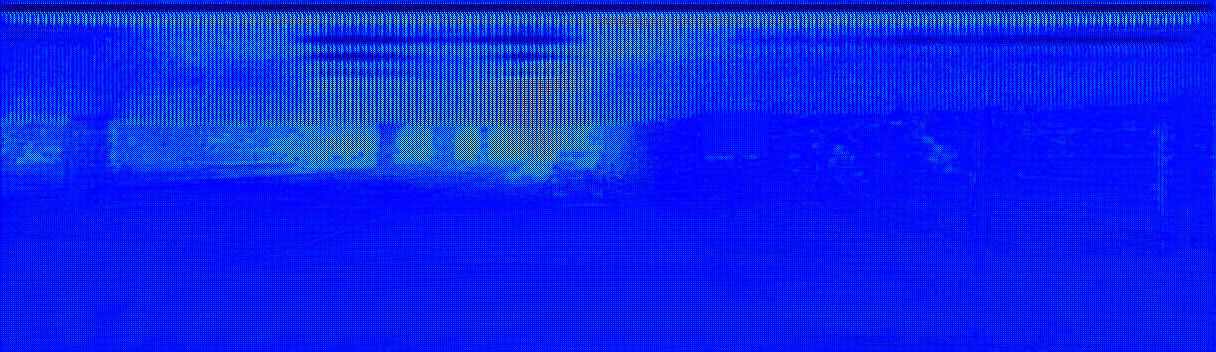}

&
\includegraphics[valign=m,width=0.33\textwidth]{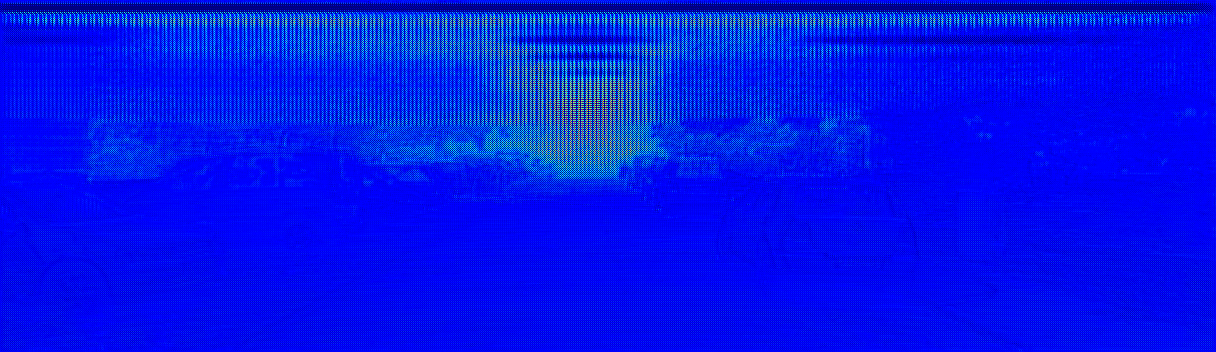}
\\
\vspace{2mm}

\rotatebox[origin=c]{90}{SG-Dep.}
&
\includegraphics[valign=m,width=0.33\textwidth]{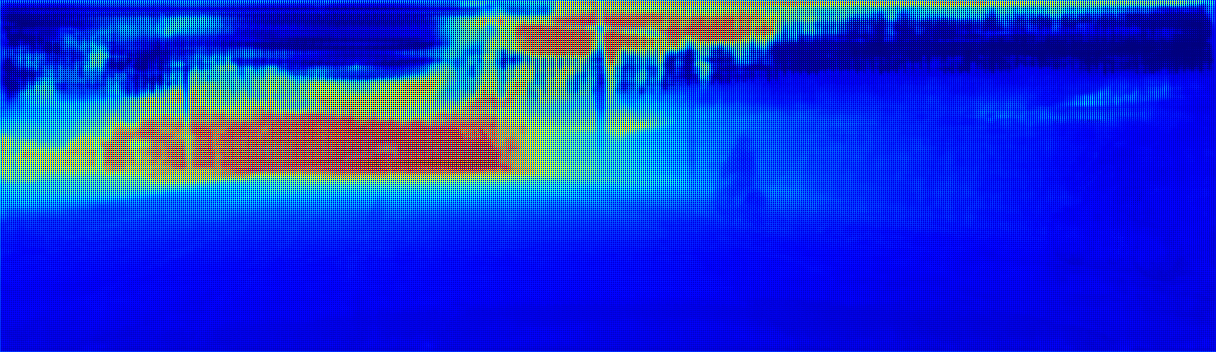}

&
\includegraphics[valign=m,width=0.33\textwidth]{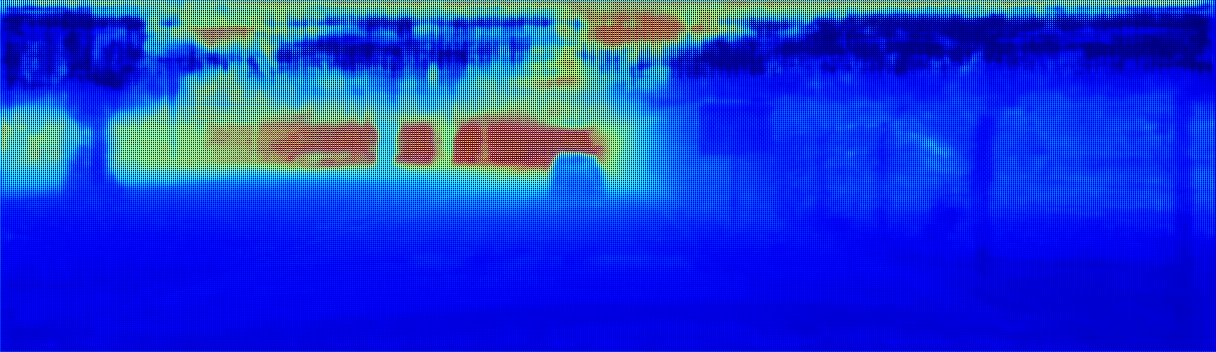}

&
\includegraphics[valign=m,width=0.33\textwidth]{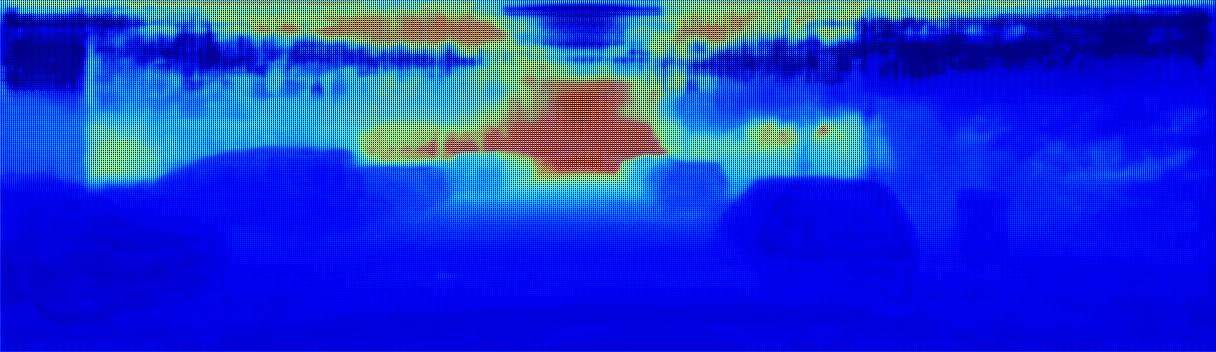}
\\
\vspace{2mm}

\rotatebox[origin=c]{90}{DG-Dep.}
&
\includegraphics[valign=m,width=0.33\textwidth]{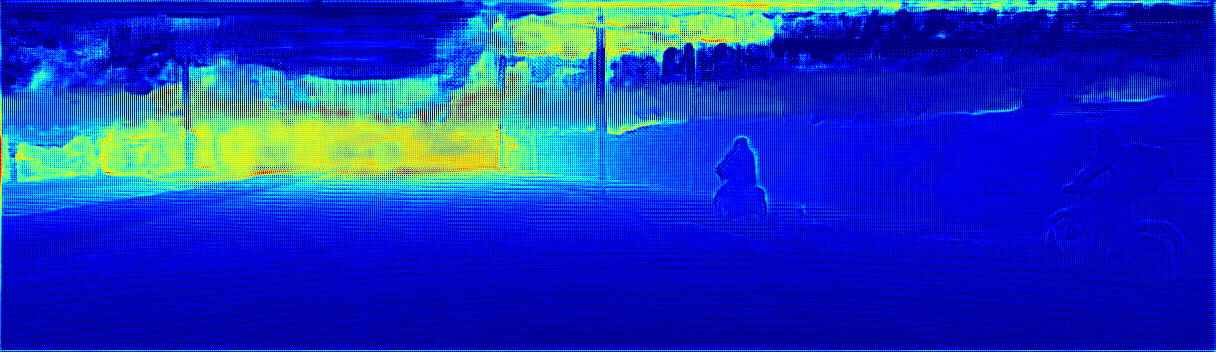}

&
\includegraphics[valign=m,width=0.33\textwidth]{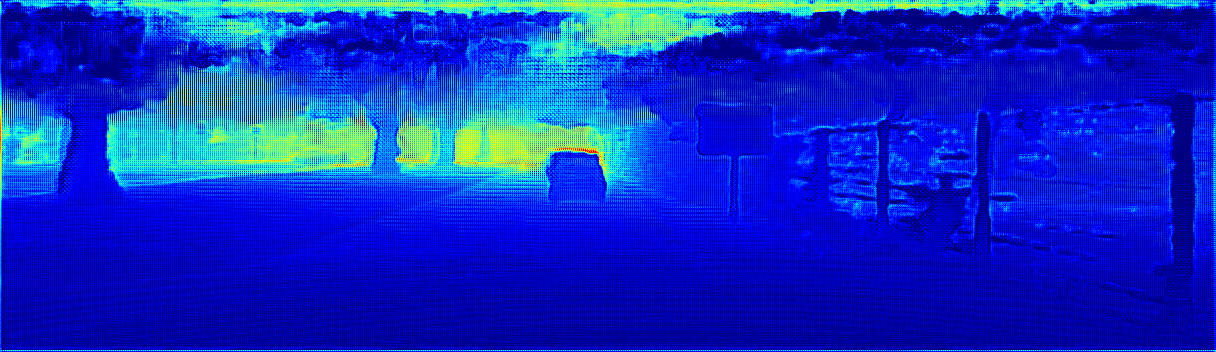}

&
\includegraphics[valign=m,width=0.33\textwidth]{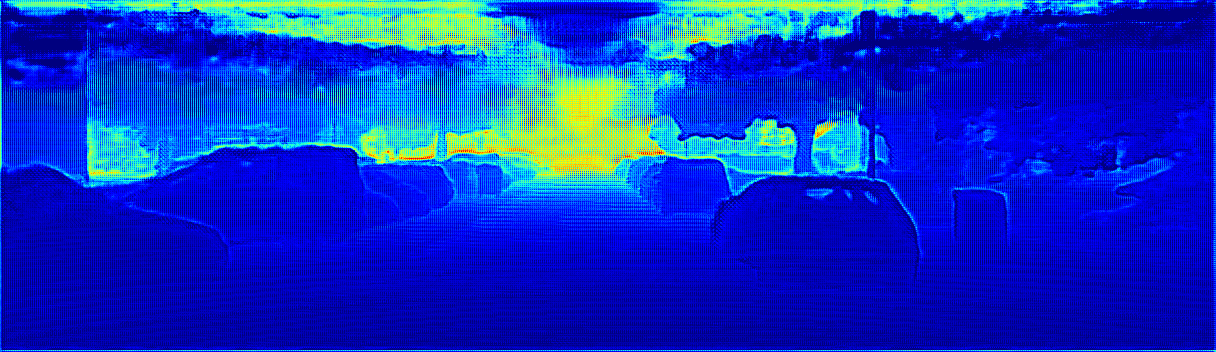}
\\
\vspace{2mm}

\rotatebox[origin=c]{90}{Fused}
&
\includegraphics[valign=m,width=0.33\textwidth]{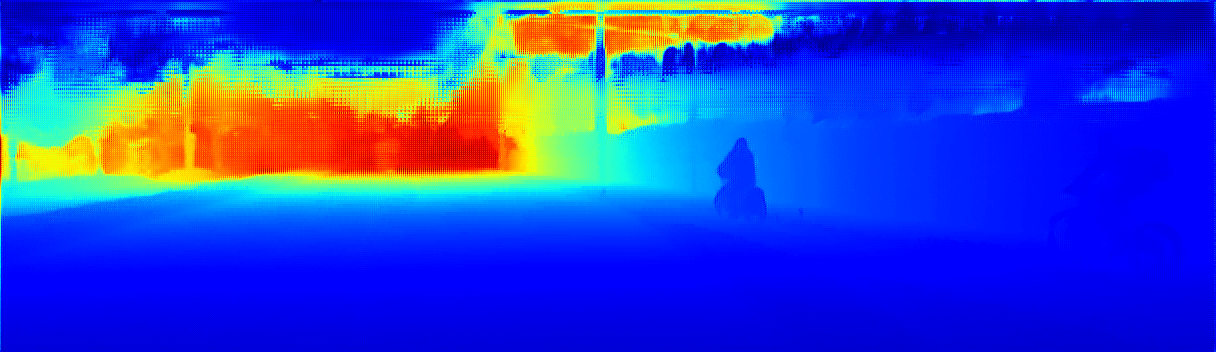}

&
\includegraphics[valign=m,width=0.33\textwidth]{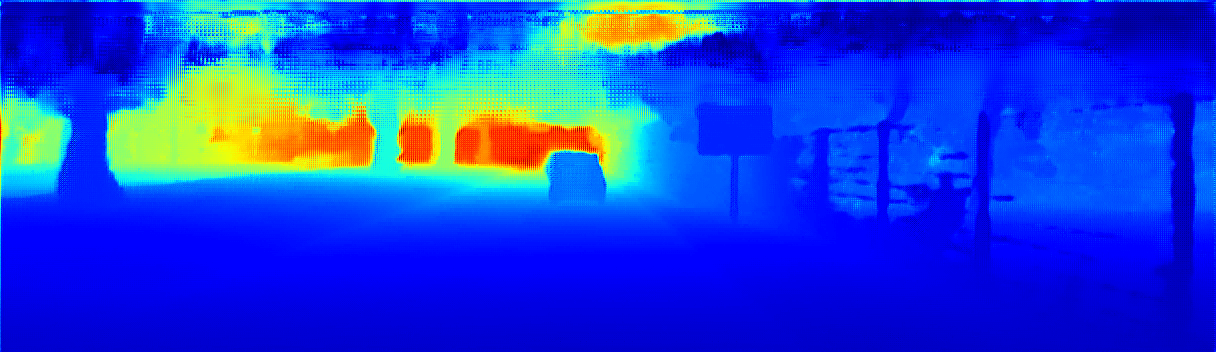}

&
\includegraphics[valign=m,width=0.33\textwidth]{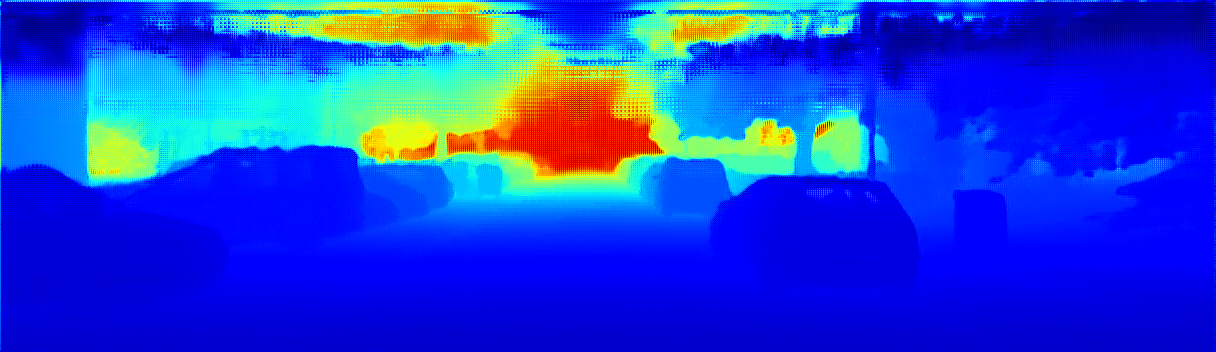}
\\
\vspace{2mm}

\rotatebox[origin=c]{90}{Refined}
&
\includegraphics[valign=m,width=0.33\textwidth]{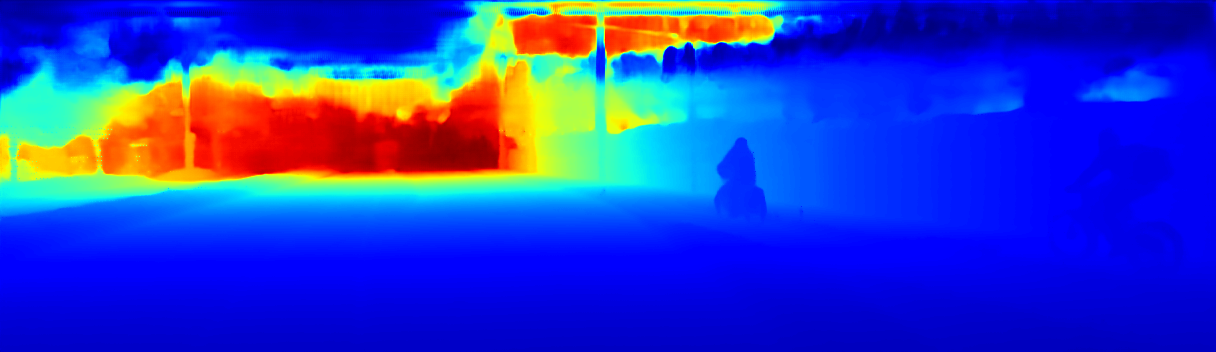}

&
\includegraphics[valign=m,width=0.33\textwidth]{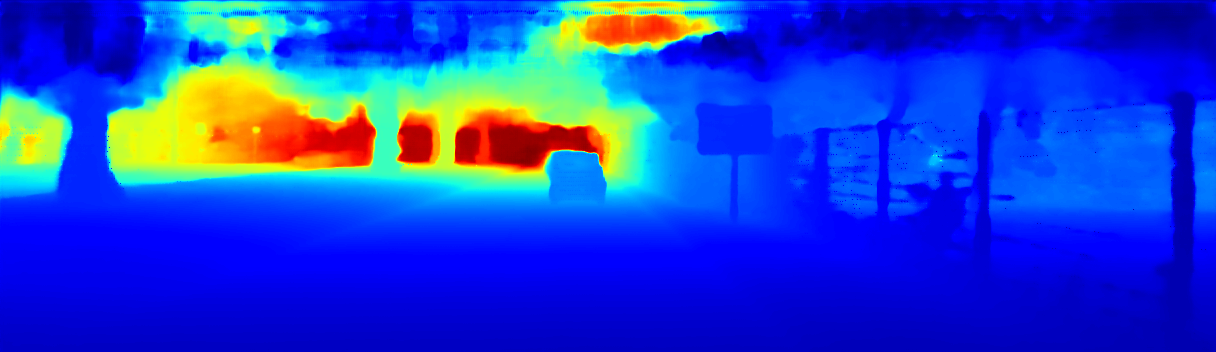}

&
\includegraphics[valign=m,width=0.33\textwidth]{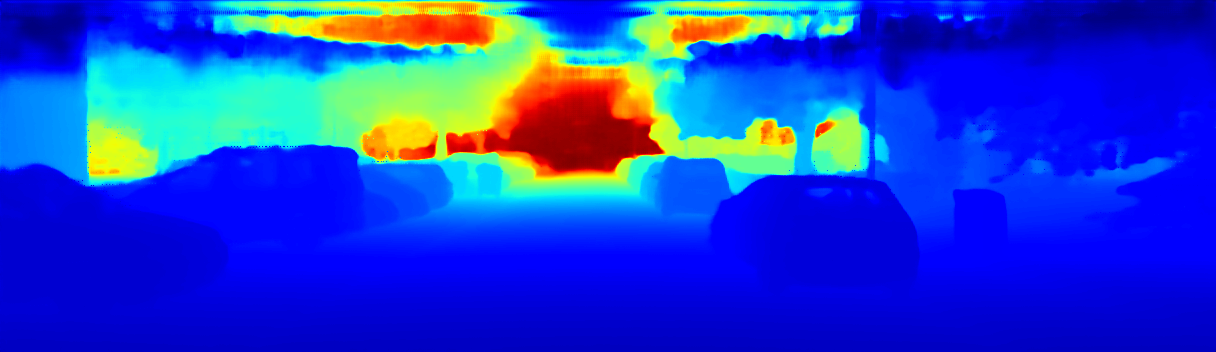}

\end{tabular}
\caption{Intermediate depths produced by each branch of SemAttNet on KITTI validation dataset \cite{KITTI, kitti_2}.~SD stands for LiDAR sparse depth, GT for sparse ground truth, CG-Dep. for color-guided dense depth, SG-Dep. for semantic-guided dense depth, and DG-Dep. for depth-guided dense depth.~The depth produced by the fusion of CG, SG and DG branches outputs is represented by Fused label.~The fused depth is passed to CSPN++ with Atrous convolutions for further refinement and it is represented by Refined label.}
\label{three_branch_results}
\end{figure*}

We used PyTorch \cite{PyTorch} framework to implement the three-branch backbone and CSPN++ with Atrous convolutions.~The three-branch backbone is trained on two NVIDIA A100 GPUs, whereas the full model is trained on three NVIDIA RTXA6000 GPUs.~Both models use same setting of ADAM optimizer \cite{kingma2017adam} with $\beta_1$ of  0.9, $\beta_1$ of 0.99 and weight decay of $10^{-6}$.~For efficient learning, we performed data prepossessing methods including random crop, flip and color jitter \cite{ma2018sparse} on the dataset.~The random crops are performed at $320 \times 1216$ resolution. 

 The three-branch backbone is trained with a batch size of 8 and an initial learning rate of 0.00128. The initial learning rate is decayed using ReduceLROnPlateu scheduler.~Furthermore, we assign an initial weight of 0.2 to  $\lambda_{cg}$, $\lambda_{sg}$ and $\lambda_{dg}$ coefficients, but we decay the weight to 0 after few epochs. We train the backbone for 60 epochs. To train the CSPN++ with Atrous convolutions, similar to PENet \cite{hu2020PENet}, we first perform the initialization step and then fully train the models for 95 epochs.

\subsection{Evaluation on KITTI dataset}
Table \ref{table1} shows the quantitative comparison of our method with other state-of-the-art methods.~Our proposed method, SemAttNet, outperforms the previous state-of-the-art methods under the primary evaluation metric RMSE at the time of submission, and presents comparable performance on other evaluation metrics.~Specifically, in contrast to previous state-of-the-arts, i.e., RigNet and PENet, we observe a difference of $4.03mm$ and $21.03mm$ to the performance of SemAttNet.

Furthermore, Figure \ref{qualitative_results} presents the qualitative results of our approach on KITTI \cite{KITTI, kitti_2} depth completion test set.~The results of all of the methods are obtained from the online KITTI depth completion leaderboard.~Compared to current state-of-the-art methods, our predicted dense depth maps are precise and consistent in the areas suffering from variance in optical changes such as shadows and reflections.~For instance, in row 5 of Figure \ref{qualitative_results}, the error map shows that our proposed method can predict accurate depth values irrespective of the reflections at car windows.~Similarly, in row 5, in comparison to other methods, our approach accurately predicts the structure of the person and the car due to the semantic understanding of the environment.~The other methods over-segment both human and car, leading to a high error in the error map.


\subsection{Ablation Studies}
In this section, we first investigate the intermediate produced by each step of our depth completion process and then discuss the impact of each component proposed by our approach on final performance.~All of the ablation studies are performed on selected KITTI validation dataset.~Table \ref{table2} depicts the results of our experiments with different settings.~Note, that we consider RMSE as our primary metric to rank the performance of methods.

\subsubsection{Analysis of Three-branch Backbone}

Figure \ref{three_branch_results} shows the intermediate depths produced by each branch.~The color depth is represented by CG-Dep., semantic depth by SG-Dep., guided dense depth by DG-Dep. in rows 5, 6, and 7, respectively.~Furthermore rows 8 and 9 represents the fused and refined dense depth maps.~The color depth is an initial dense estimate of the sparse depth.~It is produced by color branch and the objective of the color branch is to learn color dominant features of the scene, which are very important for semantic-guided and depth-guided branches.~Similarly, semantic-guided (SG) branch outputs semantic depth and learns to discriminates between different scene objects such as cycles, trees, towers, walls, etc.~However, semantic depth is not precise and consists of blurry structures.~To tackle this problem, we utilize a depth-guided (DG) branch, which generates guided dense depth map with sharp structures, especially at the object boundaries.

The dense depth maps produced by color-guided, semantic-guided, and depth-guided branches are adaptively fused to combine the multi-modal information learned by each branch.~The fused depth map shows that it contains depth information from all of the branches.~Finally, the fused depth map is sent to the CSPN++ module with Atrous convolutions \cite{cheng2020cspn++,hu2020PENet} for further refinement.~CSPN++ module also helps recover depth values at valid pixels of sparse depths in the final dense depth map.

\subsubsection{Effectiveness of SG branch}

From methods (b) and (d) listed in Table \ref{table2}, we can observe that by using the proposed semantic-guided branch even with naive concatenation method for fusion, there is a significant improvement of $26.5mm$ in the RMSE metric.~This shows the promise and importance of our proposed SG branch.~It also proves our claim that semantic cues are important for dense depth completion.  

\subsubsection{Effectiveness of SAMMAFB}

The methods (d) and (e) in Table \ref{table2} list the performance of the naive fusion method, and SAMMAFB.~It is quite evident that SAMMAFB, i.e., method (d), outperforms the naive fusion method by achieving $2.14mm$ less error than method (f).~Moreover, if we consider methods (f) and (g) which employs CSPN++ refinement along with naive and SAMMAFB fusion methods, we see a significant improvement of $11.93mm$ of RMSE in the method utilizing the SAMMAFB for fusion.~These improvement backs our claim that SAMMAFB is a better fusion strategy than naive fusion methods.

\subsubsection{Effectiveness of CSPN++ refinement}

Inspired by original CSPN++ \cite{cheng2020cspn++}, we also use 12 iterations for neighborhood propagation.~However, similar to PENet \cite{hu2020PENet}, we apply a dilation strategy to the original CSPN++ in which for the first 6 iterations, we use a dilation rate of 2, and then we decrease it to 1 for remaining iterations to propagate the neighborhood of each pixel.~From Table \ref{table2}, the method (g) that employs CSPN++ refinement with the dilation strategy, outperforms method (f) with a significant margin of $14.89mm$ in RMSE metric.~This proves that CSPN++ is an effective method for dense depth refinement.

\section{Conclusion}
In this paper, we presented a novel three-branch backbone consisting of color-guided, semantic-guided, and depth-guided branches for dense depth completion.~Existing image-guided methods focus only on RGB images to extract both color and semantic cues of the scene. However, RGB images suffer from artifacts such as varying contrast and shadows. To tackle this problem, in our three-branch backbone, we introduced a novel semantic-guided branch, which, irrespective of any optical changes in RGB images, learns to understand the semantics of the scene and reduces the reliance on RGB images.~Moreover, instead of fusing features naively through addition or concatenation, we also utilized semantic-aware multi-modal fusion block (SAMMAFB) for feature fusion between all branches.~In addition, we also implement CSPN++ with Atrous convolutions to further refine the dense depth maps.~Extensive experiments show that our results are superior both quantitatively and qualitatively compared to previous state-of-the-art methods.

{\small
\bibliographystyle{IEEEtran}
\bibliography{access}

@String(IJCV = {Int. J. Comput. Vis.})

@String(CVPR= {IEEE Conf. Comput. Vis. Pattern Recog.})

@String(ICCV= {Int. Conf. Comput. Vis.})

@String(ECCV= {Eur. Conf. Comput. Vis.})

@String(TOG= {ACM Trans. Graph.})

@String(ICIP = {IEEE Int. Conf. Image Process.})

@String(ACCV  = {ACCV})

@String(ICLR = {Int. Conf. Learn. Represent.})

@String(AAAI = {AAAI})

@String(IJCV  = {IJCV})

@String(CVPR  = {CVPR})

@String(ICCV  = {ICCV})

@String(ECCV  = {ECCV})

@String(TOG   = {ACM TOG})

@String(ICIP  = {ICIP})

@String(ICLR  = {ICLR})

@ARTICLE{DepthNet,
  author={Bai, Lin and Zhao, Yiming and Elhousni, Mahdi and Huang, Xinming},
  journal={IEEE Access}, 
  title={DepthNet: Real-Time LiDAR Point Cloud Depth Completion for Autonomous Vehicles}, 
  year={2020},
  volume={8},
  number={},
  pages={227825-227833},
  doi={10.1109/ACCESS.2020.3045681}}

@inproceedings{park2020non,
  title={Non-local spatial propagation network for depth completion},
  author={Park, Jinsun and Joo, Kyungdon and Hu, Zhe and Liu, Chi-Kuei and So Kweon, In},
  booktitle={European Conference on Computer Vision (ECCV)},
  pages={120--136},
  year={2020},
  organization={Springer}
}

@INPROCEEDINGS{augmented_app,  author={Kalia, M. and Navab, N. and Salcudean, T.},  booktitle={2019 International Conference on Robotics and Automation (ICRA)},   title={A Real-Time Interactive Augmented Reality Depth Estimation Technique for Surgical Robotics},   year={2019},  volume={},  number={},  pages={8291-8297},  doi={10.1109/ICRA.2019.8793610}}

@article{holynski2018fast,
  title={Fast depth densification for occlusion-aware augmented reality},
  author={Holynski, Aleksander and Kopf, Johannes},
  journal={ACM Transactions on Graphics (ToG)},
  volume={37},
  number={6},
  pages={1--11},
  year={2018},
  publisher={ACM New York, NY, USA}
}

@ARTICLE{3DReconstruction,
  author={Nguyen, Trong-Nguyen and Huynh, Huu-Hung and Meunier, Jean},
  journal={IEEE Access}, 
  title={3D Reconstruction With Time-of-Flight Depth Camera and Multiple Mirrors}, 
  year={2018},
  volume={6},
  number={},
  pages={38106-38114},
  doi={10.1109/ACCESS.2018.2854262}}

@INPROCEEDINGS{fisheyeautonomous,
  author={Cui, Zhaopeng and Heng, Lionel and Yeo, Ye Chuan and Geiger, Andreas and Pollefeys, Marc and Sattler, Torsten},
  booktitle={2019 International Conference on Robotics and Automation (ICRA)}, 
  title={Real-Time Dense Mapping for Self-Driving Vehicles using Fisheye Cameras}, 
  year={2019},
  volume={},
  number={},
  pages={6087-6093},
  doi={10.1109/ICRA.2019.8793884}}

@Article{obstacle_avoid,
AUTHOR = {Huang, Hsieh-Chang and Hsieh, Ching-Tang and Yeh, Cheng-Hsiang},
TITLE = {An Indoor Obstacle Detection System Using Depth Information and Region Growth},
JOURNAL = {Sensors},
VOLUME = {15},
YEAR = {2015},
NUMBER = {10},
PAGES = {27116--27141},
URL = {https://www.mdpi.com/1424-8220/15/10/27116},
PubMedID = {26512674},
ISSN = {1424-8220},
DOI = {10.3390/s151027116}
}

@article{tang2020learning,
  title={Learning guided convolutional network for depth completion},
  author={Tang, Jie and Tian, Fei-Peng and Feng, Wei and Li, Jian and Tan, Ping},
  journal={IEEE Transactions on Image Processing},
  volume={30},
  pages={1116--1129},
  year={2020},
  publisher={IEEE}
}

@inproceedings{KITTI,
  title={Sparsity invariant cnns},
  author={Uhrig, Jonas and Schneider, Nick and Schneider, Lukas and Franke, Uwe and Brox, Thomas and Geiger, Andreas},
  booktitle={2017 international conference on 3D Vision (3DV)},
  pages={11--20},
  year={2017},
  organization={IEEE}
}

@InProceedings{Qiu_2019_CVPR,
author = {Qiu, Jiaxiong and Cui, Zhaopeng and Zhang, Yinda and Zhang, Xingdi and Liu, Shuaicheng and Zeng, Bing and Pollefeys, Marc},
title = {DeepLiDAR: Deep Surface Normal Guided Depth Prediction for Outdoor Scene From Sparse LiDAR Data and Single Color Image},
booktitle = {The IEEE Conference on Computer Vision and Pattern Recognition (CVPR)},
month = {June},
year = {2019}
}

@inproceedings{hu2020PENet,
  title={Penet: Towards precise and efficient image guided depth completion},
  author={Hu, Mu and Wang, Shuling and Li, Bin and Ning, Shiyu and Fan, Li and Gong, Xiaojin},
  booktitle={2021 IEEE International Conference on Robotics and Automation (ICRA)},
  pages={13656--13662},
  year={2021},
  organization={IEEE}
}

@inproceedings{lee2021depth,
  title={Depth Completion using Plane-Residual Representation},
  author={Lee, Byeong-Uk and Lee, Kyunghyun and Kweon, In So},
  booktitle={Proceedings of the IEEE/CVF Conference on Computer Vision and Pattern Recognition (CVPR)},
  pages={13916--13925},
  year={2021}
}

@article{yan2021rignet,
  title={RigNet: Repetitive image guided network for depth completion},
  author={Yan, Zhiqiang and Wang, Kun and Li, Xiang and Zhang, Zhenyu and Xu, Baobei and Li, Jun and Yang, Jian},
  journal={arXiv preprint arXiv:2107.13802},
  year={2021}
}

@inproceedings{tof,
author = {Gavriel J. Iddan and Giora Yahav},
title = {{Three-dimensional imaging in the studio and elsewhere}},
volume = {4298},
booktitle = {Three-Dimensional Image Capture and Applications IV},
editor = {Brian D. Corner and Joseph H. Nurre and Roy P. Pargas},
organization = {International Society for Optics and Photonics},
publisher = {SPIE},
pages = {48 -- 55},
year = {2001},
doi = {10.1117/12.424913},
URL = {https://doi.org/10.1117/12.424913}
}

@ARTICLE{Multi-TaskGan,  author={Zhang, Chongzhen and Tang, Yang and Zhao, Chaoqiang and Sun, Qiyu and Ye, Zhencheng and Kurths, Jürgen},  journal={IEEE Transactions on Neural Networks and Learning Systems},   title={Multitask GANs for Semantic Segmentation and Depth Completion With Cycle Consistency},   year={2021},  volume={},  number={},  pages={1-12},  doi={10.1109/TNNLS.2021.3072883}}

@article{zhao2021adaptive,
  title={Adaptive context-aware multi-modal network for depth completion},
  author={Zhao, Shanshan and Gong, Mingming and Fu, Huan and Tao, Dacheng},
  journal={IEEE Transactions on Image Processing},
  volume={30},
  pages={5264--5276},
  year={2021},
  publisher={IEEE}
}

@ARTICLE{Alhaija2018IJCV,
  author = {Hassan Alhaija and Siva Mustikovela and Lars Mescheder and Andreas Geiger and Carsten Rother},
  title = {Augmented Reality Meets Computer Vision: Efficient Data Generation for Urban Driving Scenes},
  journal = {International Journal of Computer Vision (IJCV)},
  year = {2018}
}

@article{wu2016wider,
  title={Wider or deeper: Revisiting the resnet model for visual recognition},
  author={Wu, Zifeng and Shen, Chunhua and Van Den Hengel, Anton},
  journal={Pattern Recognition},
  volume={90},
  pages={119--133},
  year={2019},
  publisher={Elsevier}
}

@inproceedings{xu2020deformable,
  title={Deformable spatial propagation networks for depth completion},
  author={Xu, Zheyuan and Yin, Hongche and Yao, Jian},
  booktitle={2020 IEEE International Conference on Image Processing (ICIP)},
  pages={913--917},
  year={2020},
  organization={IEEE}
}

@inproceedings{yu2016multiscale,
  author    = {Fisher Yu and
               Vladlen Koltun},
  editor    = {Yoshua Bengio and
               Yann LeCun},
  title     = {Multi-Scale Context Aggregation by Dilated Convolutions},
  booktitle = {4th International Conference on Learning Representations, {ICLR} 2016,
               San Juan, Puerto Rico, May 2-4, 2016, Conference Track Proceedings},
  year      = {2016},
  url       = {http://arxiv.org/abs/1511.07122},
  bibsource = {dblp computer science bibliography, https://dblp.org}
}

@inproceedings{liu2021fcfr,
  title={FCFR-Net: Feature Fusion based Coarse-to-Fine Residual Learning for Depth Completion},
  author={Liu, Lina and Song, Xibin and Lyu, Xiaoyang and Diao, Junwei and Wang, Mengmeng and Liu, Yong and Zhang, Liangjun},
  booktitle={Proceedings of the AAAI Conference on Artificial Intelligence},
  volume={35},
  number={3},
  pages={2136--2144},
  year={2021}
}

@inproceedings{cheng2020cspn++,
  title={Cspn++: Learning context and resource aware convolutional spatial propagation networks for depth completion},
  author={Cheng, Xinjing and Wang, Peng and Guan, Chenye and Yang, Ruigang},
  booktitle={Proceedings of the AAAI Conference on Artificial Intelligence},
  volume={34},
  number={07},
  pages={10615--10622},
  year={2020}
}

@inproceedings{chen2019learning,
  title={Learning joint 2d-3d representations for depth completion},
  author={Chen, Yun and Yang, Bin and Liang, Ming and Urtasun, Raquel},
  booktitle={Proceedings of the IEEE/CVF International Conference on Computer Vision (ICCV)},
  pages={10023--10032},
  year={2019}
}

@inproceedings{chen2019towards,
  title={Towards scene understanding: Unsupervised monocular depth estimation with semantic-aware representation},
  author={Chen, Po-Yi and Liu, Alexander H and Liu, Yen-Cheng and Wang, Yu-Chiang Frank},
  booktitle={Proceedings of the IEEE/CVF Conference on Computer Vision and Pattern Recognition (CVPR)},
  pages={2624--2632},
  year={2019}
}

@inproceedings{schneider2016semantically,
  title={Semantically guided depth upsampling},
  author={Schneider, Nick and Schneider, Lukas and Pinggera, Peter and Franke, Uwe and Pollefeys, Marc and Stiller, Christoph},
  booktitle={German conference on pattern recognition (GCPR)},
  pages={37--48},
  year={2016},
  organization={Springer}
}

@inproceedings{vangansbeke2019sparse,
  title={Sparse and noisy lidar completion with rgb guidance and uncertainty},
  author={Van Gansbeke, Wouter and Neven, Davy and De Brabandere, Bert and Van Gool, Luc},
  booktitle={2019 16th international conference on machine vision applications (MVA)},
  pages={1--6},
  year={2019},
  organization={IEEE}
}

@article{cheng2018depth,
  title={Learning depth with convolutional spatial propagation network},
  author={Cheng, Xinjing and Wang, Peng and Yang, Ruigang},
  journal={IEEE transactions on pattern analysis and machine intelligence (TPAMI)},
  volume={42},
  number={10},
  pages={2361--2379},
  year={2019},
  publisher={IEEE}
}

@article{bai2020depthnet,
  title={DepthNet: Real-Time LiDAR Point Cloud Depth Completion for Autonomous Vehicles},
  author={Bai, Lin and Zhao, Yiming and Elhousni, Mahdi and Huang, Xinming},
  journal={IEEE Access},
  volume={8},
  pages={227825--227833},
  year={2020},
  publisher={IEEE}
}

@inproceedings{eldesokey2020uncertainty,
  title={Uncertainty-aware cnns for depth completion: Uncertainty from beginning to end},
  author={Eldesokey, Abdelrahman and Felsberg, Michael and Holmquist, Karl and Persson, Michael},
  booktitle={Proceedings of the IEEE/CVF Conference on Computer Vision and Pattern Recognition (CVPR)},
  pages={12014--12023},
  year={2020}
}

@inproceedings{chodosh2018deep,
  title={Deep convolutional compressed sensing for lidar depth completion},
  author={Chodosh, Nathaniel and Wang, Chaoyang and Lucey, Simon},
  booktitle={Asian Conference on Computer Vision (ACCV)},
  pages={499--513},
  year={2018},
  organization={Springer}
}

@article{gu2021denselidar,
  title={DenseLiDAR: A Real-Time Pseudo Dense Depth Guided Depth Completion Network},
  author={Gu, Jiaqi and Xiang, Zhiyu and Ye, Yuwen and Wang, Lingxuan},
  journal={IEEE Robotics and Automation Letters},
  volume={6},
  number={2},
  pages={1808--1815},
  year={2021},
  publisher={IEEE}
}

@article{liu2017learning,
  title={Learning affinity via spatial propagation networks},
  author={Liu, Sifei and De Mello, Shalini and Gu, Jinwei and Zhong, Guangyu and Yang, Ming-Hsuan and Kautz, Jan},
  journal={arXiv preprint arXiv:1710.01020},
  year={2017}
}

@inproceedings{dspn,
  title={Deformable spatial propagation networks for depth completion},
  author={Xu, Zheyuan and Yin, Hongche and Yao, Jian},
  booktitle={2020 IEEE International Conference on Image Processing (ICIP)},
  pages={913--917},
  year={2020},
  organization={IEEE}
}

@article{liou2014autoencoder,
  title={Autoencoder for words},
  author={Liou, Cheng-Yuan and Cheng, Wei-Chen and Liou, Jiun-Wei and Liou, Daw-Ran},
  journal={Neurocomputing},
  volume={139},
  pages={84--96},
  year={2014},
  publisher={Elsevier}
}

@inproceedings{he2016deep,
  title={Deep residual learning for image recognition},
  author={He, Kaiming and Zhang, Xiangyu and Ren, Shaoqing and Sun, Jian},
  booktitle={Proceedings of the IEEE conference on computer vision and pattern recognition (CVPR)},
  pages={770--778},
  year={2016}
}

@INPROCEEDINGS{semantic,  author={Lu, Kaiyue and Barnes, Nick and Anwar, Saeed and Zheng, Liang},  booktitle={2020 IEEE/CVF Conference on Computer Vision and Pattern Recognition (CVPR)},   title={From Depth What Can You See? Depth Completion via Auxiliary Image Reconstruction},   year={2020},  volume={},  number={},  pages={11303-11312},  doi={10.1109/CVPR42600.2020.01132}}

@INPROCEEDINGS{Kitti_2,  author={Geiger, Andreas and Lenz, Philip and Urtasun, Raquel},  booktitle={2012 IEEE Conference on Computer Vision and Pattern Recognition},   title={Are we ready for autonomous driving? The KITTI vision benchmark suite},   year={2012},  volume={},  number={},  pages={3354-3361},  doi={10.1109/CVPR.2012.6248074}}

@inproceedings{woo2018cbam,
  title={Cbam: Convolutional block attention module},
  author={Woo, Sanghyun and Park, Jongchan and Lee, Joon-Young and Kweon, In So},
  booktitle={Proceedings of the European conference on computer vision (ECCV)},
  pages={3--19},
  year={2018}
}

@article{fooladgar2019multi,
  title={Multi-Modal Attention-based Fusion Model for Semantic Segmentation of RGB-Depth Images},
  author={Fooladgar, Fahimeh and Kasaei, Shohreh},
  journal={arXiv preprint arXiv:1912.11691},
  year={2019}
}

@article{chen2017deeplab,
  title={Deeplab: Semantic image segmentation with deep convolutional nets, atrous convolution, and fully connected crfs},
  author={Chen, Liang-Chieh and Papandreou, George and Kokkinos, Iasonas and Murphy, Kevin and Yuille, Alan L},
  journal={IEEE transactions on pattern analysis and machine intelligence (TPAMI)},
  volume={40},
  number={4},
  pages={834--848},
  year={2017},
  publisher={IEEE}
}

@inproceedings{hori2017attention,
  title={Attention-based multimodal fusion for video description},
  author={Hori, Chiori and Hori, Takaaki and Lee, Teng-Yok and Zhang, Ziming and Harsham, Bret and Hershey, John R and Marks, Tim K and Sumi, Kazuhiko},
  booktitle={Proceedings of the IEEE international conference on computer vision (ICCV)},
  pages={4193--4202},
  year={2017}
}

@inproceedings{li2020attention,
  title={Attention-Based Multimodal Fusion for Estimating Human Emotion in Real-World HRI},
  author={Li, Yuanchao and Zhao, Tianyu and Shen, Xun},
  booktitle={Companion of the 2020 ACM/IEEE International Conference on Human-Robot Interaction},
  pages={340--342},
  year={2020}
}

@article{huddar2021attention,
  title={Attention-based multimodal contextual fusion for sentiment and emotion classification using bidirectional LSTM},
  author={Huddar, Mahesh G and Sannakki, Sanjeev S and Rajpurohit, Vijay S},
  journal={Multimedia Tools and Applications},
  volume={80},
  number={9},
  pages={13059--13076},
  year={2021},
  publisher={Springer}
}

@incollection{PyTorch,
title = {PyTorch: An Imperative Style, High-Performance Deep Learning Library},
author = {Paszke, Adam and Gross, Sam and Massa, Francisco and Lerer, Adam and Bradbury, James and Chanan, Gregory and Killeen, Trevor and Lin, Zeming and Gimelshein, Natalia and Antiga, Luca and Desmaison, Alban and Kopf, Andreas and Yang, Edward and DeVito, Zachary and Raison, Martin and Tejani, Alykhan and Chilamkurthy, Sasank and Steiner, Benoit and Fang, Lu and Bai, Junjie and Chintala, Soumith},
booktitle = {Advances in Neural Information Processing Systems 32},
editor = {H. Wallach and H. Larochelle and A. Beygelzimer and F. d\textquotesingle Alch\'{e}-Buc and E. Fox and R. Garnett},
pages = {8024--8035},
year = {2019},
publisher = {Curran Associates, Inc.},
url = {http://papers.neurips.cc/paper/9015-pytorch-an-imperative-style-high-performance-deep-learning-library.pdf}
}

@article{kingma2017adam,
author = {Kingma, Diederik and Ba, Jimmy},
year = {2014},
month = {12},
pages = {},
title = {Adam: A Method for Stochastic Optimization},
journal = {International Conference on Learning Representations (ICLR)}
}

@inproceedings{ma2018sparse,
  title={Sparse-to-dense: Depth prediction from sparse depth samples and a single image},
  author={Ma, Fangchang and Karaman, Sertac},
  booktitle={2018 IEEE international conference on robotics and automation (ICRA)},
  pages={4796--4803},
  year={2018},
  organization={IEEE}
}

@inproceedings{imran2021depth,
  title={Depth Completion with Twin Surface Extrapolation at Occlusion Boundaries},
  author={Imran, Saif and Liu, Xiaoming and Morris, Daniel},
  booktitle={Proceedings of the IEEE/CVF Conference on Computer Vision and Pattern Recognition (CVPR)},
  pages={2583--2592},
  year={2021}
}

@inproceedings{ma2019self,
  title={Self-supervised sparse-to-dense: Self-supervised depth completion from lidar and monocular camera},
  author={Ma, Fangchang and Cavalheiro, Guilherme Venturelli and Karaman, Sertac},
  booktitle={2019 International Conference on Robotics and Automation (ICRA)},
  pages={3288--3295},
  year={2019},
  organization={IEEE}
}
}
\vskip 0pt plus -1fil
\begin{IEEEbiography}[{\includegraphics[width=1in,height=1.25in,clip,keepaspectratio]{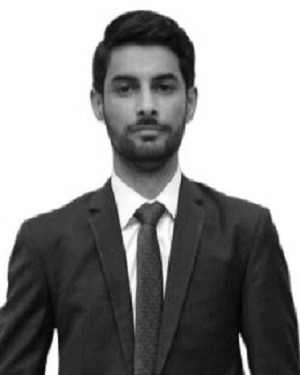}}]{Danish Nazir} received the bachelor's degree in computer science from Commission on Science and Technology for Sustainable Development in the South (COMSATS)
University Islamabad, Pakistan, in 2016, and currently pursuing  M.S. degree with core specialization in intelligent systems from the Technical University of Kaiserslautern.
He is also working as a research assistant at German Research Center for Artificial Intelligence (DFKI GmbH) under the supervision of Dr. Muhammad Zeshan Afzal. He in working in the field of Autonomous Driving Assistance Systems (ADAS) and his research interests include semantic scene understanding and sensor fusion between multi-modal information.

\end{IEEEbiography}
\vskip 10pt plus -30fil
\begin{IEEEbiography}[{\includegraphics[width=1in,height=1.25in,clip,keepaspectratio]{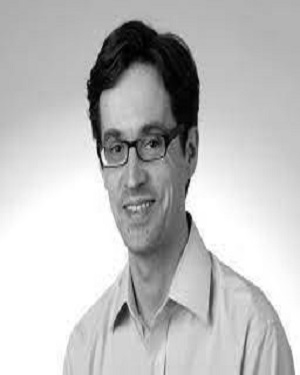}}]{Alain Pagani} studied computer science at the Ecole Centrale (Lyon, France, 2001) and at the Technical University of Darmstadt (Germany, 2003). He started his career as a researcher at the Augmented Reality department of the Fraunhofer Institute for Computer Graphics (Fraunhofer IGD, Darmstadt, Germany). In 2008, he participated to the creation of the Augmented Vision department of the German Research Center for Artificial Intelligence (DFKI) in Kaiserslautern, where he currently acts as deputy director. He obtained his PhD from the University of Kaiserslautern with a thesis entitled “Reality Models for efficient registration in Augmented Reality”, in which he tackled the problem or real-time pose estimation for AR using various mathematical models of real scenes.
His research interests include Computer Vision, Artificial Intelligence, Image Understanding, Augmented Reality and Machine Learning, and he published over 60 articles in conferences and journals. Parallel to his work at DFKI, he gives lectures at the Technical University of Kaiserslautern (“Computer Vision: Object and People Tracking”, and “3D Computer Vision”).
\end{IEEEbiography}
\vskip 0pt plus -1fil
\begin{IEEEbiography}[{\includegraphics[width=1in,height=1.25in,clip,keepaspectratio]{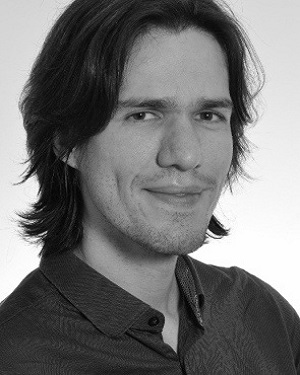}}]{Marcus Liwicki} received his M.S. degree in Computer Science from the Free University of Berlin, Germany, in 2004, his PhD degree from the University of Bern, Switzerland, in 2007, and his habilitation degree at the Technical University of Kaiserslautern, Germany, in 2011. Currently he is chaired professor at Luleå University of Technology and a senior assistant in the University of Fribourg. His research interests include machine learning, pattern recognition, artificial intelligence, human computer interaction, digital humanities, knowledge management, ubiquitous intuitive input devices, document analysis, and graph matching. He is a member of the IAPR, editor or regular reviewer for international journals, including IEEE Transactions on Pattern Analysis and Machine Intelligence, IEEE Transactions on Audio, Speech and Language Processing, International Journal of Document Analysis and Recognition (editor), Frontiers of Computer science (editor), Frontiers in Digital Humanities (editor), Pattern Recognition, and Pattern Recognition Letters. He is a member of governing board the International Graphonomics Society and a member of the International Association for Pattern Recognition where he is Vice president of the Technical Committee 6. He chaired several International Workshops on Automated Forensic Handwriting Analysis and the International Workshop on Document Analysis Systems 2014. Furthermore he serves as program committee member and reviewer of various International Conferences and workshops in the area of Computer Vision, Pattern Recognition and Document Analysis as well as Machine Learning and E-Learning.
\end{IEEEbiography}
\vskip 0pt plus -1fil
\begin{IEEEbiography}[{\includegraphics[width=1in,height=1.25in,clip,keepaspectratio]{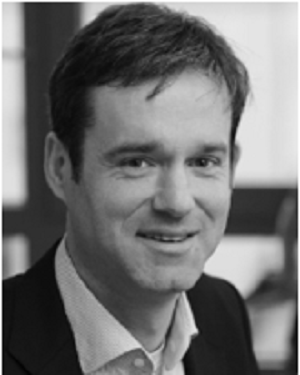}}]{Didier Stricker} is professor with the University of Kaiserslautern and scientific director with the “German Research Center for Artificial Intelligence” (DFKI) in Kaiserslautern where he leads the research department Augmented Vision. From 2002 to June 2008, he lead the department “Virtual and Augmented Reality” at the Fraunhofer Institute for Computer Graphics (Fraunhofer IGD) in Darmstadt, Germany. In this function, he initiated and participated to many national and international projects in the areas of computer vision and virtual and augmented reality. In 2006, he received the Innovation Prize of the German Society of Computer Science. He serves as reviewer for different European or national research organizations, and is a regular reviewer for the most important journals and conferences in the areas of VR/AR and computer vision.

\end{IEEEbiography}
\vskip 0pt plus -1fil
\begin{IEEEbiography}[{\includegraphics[width=1in,height=1.25in,clip,keepaspectratio]{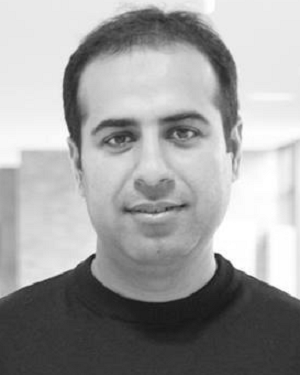}}]{Muhammad Zeshan Afzal} received his Masters degree from the University of Saarland, Germany majoring in Visual Computing in 2010 and his Ph.D. degree from the University of Technology, Kaiserslautern, Germany majoring in Artificial Intelligence in 2016. His research interests include deep learning for vision and language understanding using deep learning.  At an application level, his experience includes generic segmentation framework for natural, human activity recognition, document and, medical image analysis, scene text detection, and recognition, on-line and off-line gesture recognition. Moreover, a special interest in recurrent neural networks for sequence processing applied to images and videos. He also worked with numerics for tensor valued images. He worked both in the industry (Deep Learning and AI Lead Insiders Technologies GmbH) and academia (TU Kaiserslautern). He received the gold medal for the best graduating student in Computer Science from IUB Pakistan in 2002 and secured a DAAD(Germany) fellowship in 2007. He is a member of IAPR.
\end{IEEEbiography}

\EOD

\end{document}